\documentclass[10pt,twocolumn,letterpaper]{article}

\usepackage[pagenumbers]{iccv} %

\usepackage{graphicx}
\usepackage{amsmath}
\usepackage{amssymb}
\usepackage{booktabs}

\usepackage[nolist,nohyperlinks]{acronym}

\usepackage{array} 

\usepackage{multirow}

\usepackage{adjustbox}

\usepackage{arydshln}

\usepackage{colortbl}
\usepackage{xcolor}

\usepackage{tikz}

\definecolor{darkgreen}{RGB}{0,128,0} %

\def\fg{{\scshape fg}}
\def\bg{{\scshape bg}}
\def\full{{\scshape full}}

\def\strr{L2R$^2$}

\begin{acronym}
\acro{gt}[GT]{Ground Truth}
\acro{iou}[IoU]{Intersection over Union}
\acro{miou}[mIoU]{mean Intersection over Union}
\acro{ce}[CE]{Cross-Entropy}
\acro{vit}[ViT]{Vision Transformer}
\acro{sam}[SAM]{Segment Anything Model}
\acro{vlm}[VLM]{Vision-Language Model}
\end{acronym}

\DeclareMathOperator*{\argmax}{argmax}

\definecolor{iccvblue}{rgb}{0.21,0.49,0.74}
\usepackage[pagebackref,breaklinks,colorlinks,allcolors=iccvblue]{hyperref}

\usepackage{multirow}

\def\paperID{11476} %
\def\confName{ICCV}
\def\confYear{2025}

\title{Bringing the Context Back into Object Recognition, Robustly}

\author{Klara Janouskova \qquad Cristian Gavrus \qquad Jiri Matas \\
Visual Recognition Group, Czech Technical University in Prague\\
{\tt\footnotesize \{klara.janouskova, gavrucri, matas\}@fel.cvut.cz}
}

\begin{document}

\twocolumn[
{
\renewcommand\twocolumn[1][]{#1}%
\maketitle
\begin{center}
\setlength{\tabcolsep}{0.1pt}
    
\begin{tabular}{>{\centering\arraybackslash}m{0.195\textwidth} >{\centering\arraybackslash}m{0.195\textwidth} >{\centering\arraybackslash}m{0.195\textwidth} >{\centering\arraybackslash}m{0.195\textwidth} >{\centering\arraybackslash}m{0.195\textwidth}}

\includegraphics[width=0.19\textwidth, height=3.2cm]{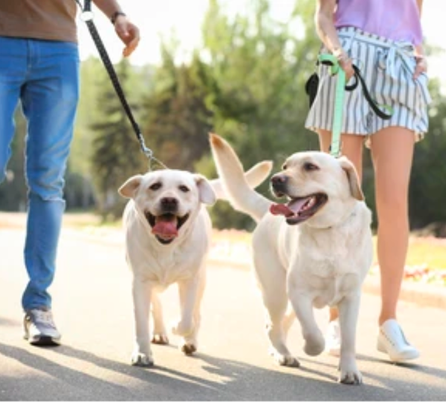}
&
\includegraphics[width=0.19\textwidth, height=3.2cm]{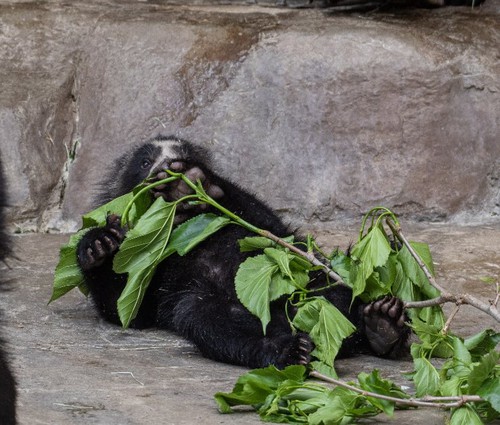}
&
    \includegraphics[width=0.19\textwidth, height=3.2cm]{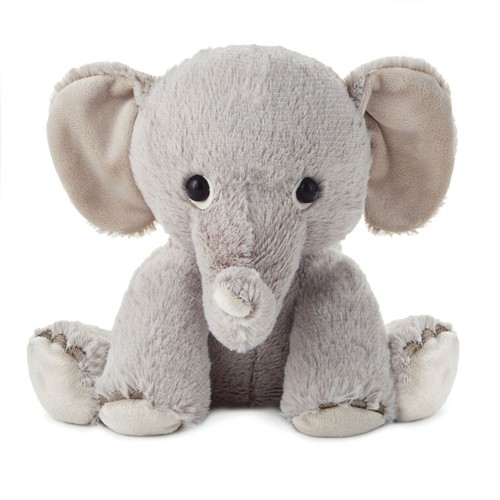}
    &
    \includegraphics[width=0.19\textwidth, height=3.2cm]{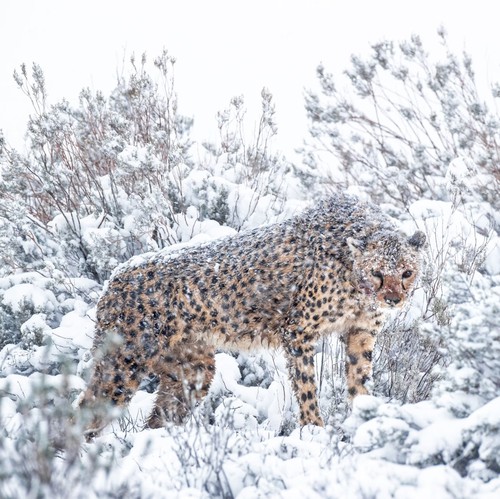} &
    \includegraphics[width=0.19\textwidth, height=3.2cm,]{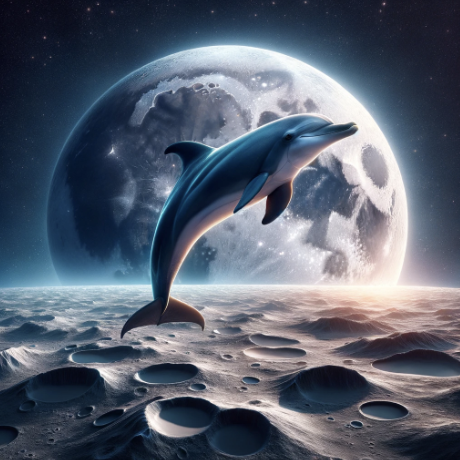} \\
    \small (a) \bg{}, the owners, critical for dog identification & 
    \small (b) the \bg{}  facilitates recognition &
    \small (c) \bg{} uninformative for classification &
    \small (d) long-tail \bg{}, not likely to appear during training
    &
    \small (e) generated \bg{} can be arbitrary 
\end{tabular}
\begin{minipage}{\textwidth}
\vspace{0.5em}
\centering
\begin{tikzpicture}
    \draw[<->, thick] (0,0) -- (0.975\textwidth,0);
    \node[fill=white, inner sep=2pt, anchor=south] at (0.1\textwidth,-0.17) {\small context critical};
    \node[fill=white, inner sep=2pt, anchor=south] at (0.87\textwidth,-0.22) {\small context misleading};
\end{tikzpicture}
\end{minipage}

    \captionof{figure}{The complementarity of foreground (\fg{}) and background (\bg{}) in recognition. The standard approach, background suppression, makes correct identification in (a) nearly impossible, and difficult in (b); the spectacled bear is the most herbivorous of all bear species. On the other hand, rare backgrounds with possibly huge diversity hurt classification -- (d) shows a cheetah after a snowfall in South Africa, not a snow leopard. In generated content (e), any \fg{} can appear on any \bg{} as in ChatGPT 4o's response to ``a dolphin on the moon".
   }
    \label{fig:teaser}
    
\end{center}

}
]

\maketitle

\begin{abstract}
In object recognition, both the subject of interest (referred to as foreground, \fg{}, for simplicity) and its surrounding context (background, \bg{}) may play an important role. 
However, standard supervised learning often leads to unintended over-reliance on the \bg{}, limiting model robustness in real-world deployment settings.
The problem is mainly addressed by suppressing the \bg{}, sacrificing context information for improved generalization.

We propose ``Localize to Recognize Robustly" (\strr{}), a novel recognition approach which exploits the benefits of context-aware classification while maintaining robustness to distribution shifts.
    \strr{} leverages advances in zero-shot detection to localize the \fg{} before recognition. 
    It improves the performance of both standard recognition with supervised training, as well as multimodal zero-shot recognition with \acs{vlm}s, while being robust to long-tail \bg{}s and distribution shifts. 
The results confirm localization before recognition is possible for a wide range of datasets and they highlight the limits of object detection on others\footnote{The code will be made publicly available on GitHub}.
\end{abstract}

\section{Introduction}
\label{sec:intro}
In standard object recognition, a neural network models the statistical distribution of the objects' appearance in the training set.
This approach has been highly successful in i.i.d. settings, particularly with moderate to large-scale training data. 

As object recognition matured, analyses of its weaknesses \cite{singla2022salient,xiao2020noise} revealed that supervised classifiers are particularly prone to unintended over-reliance on the background (\bg{}).
This seriously impacts model robustness in real-world deployment settings
as \bg{}
shortcuts \cite{geirhos2020shortcut}
perform well on training data but fail to generalize
to \bg{}s which are long-tail, i.e. rarely or never appearing in the training data, and to substantial \bg{} distribution shifts, not an uncommon situation.

Recent methods address the problem by suppression of \bg{} features. The methods can be broadly categorized into two groups: the first emphasizes \fg{} features during training \cite{bhatt2024mitigating,yang2024significant,chou2023fine,aniraj2023masking} by exploiting segmentation masks (often ground truth) or saliency maps,  the second alters the \bg{} distribution \cite{barbu2019objectnet,xiao2020noise,shetty2019not,ghosh2024aspire,wang2022clad} through image augmentation and generation techniques.

\begin{figure}[bt]
    \centering
    \setlength{\tabcolsep}{0.05pt}
    \begin{tabular}
    {>{\raggedright\arraybackslash}m{0.4\linewidth} >{\raggedright\arraybackslash}m{0.6\linewidth}}

        \includegraphics[width=0.98\linewidth]{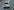} &
        \hfill
        \includegraphics[width=0.98\linewidth,trim={0cm 0.02cm 0cm 0.02cm},clip]{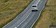} \\
        \small VLM: \hfill plug 46\% & \small \hfill  car 80\% \\
        \small \strr{} fusion: \small  \bf car & \\
         \\
         [-1em]
\includegraphics[width=0.98\linewidth]{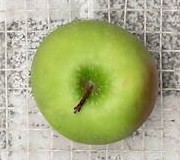} &
    \hfill    \includegraphics[width=0.98\linewidth,trim={0cm 0.4cm 0cm 0.4cm},clip]{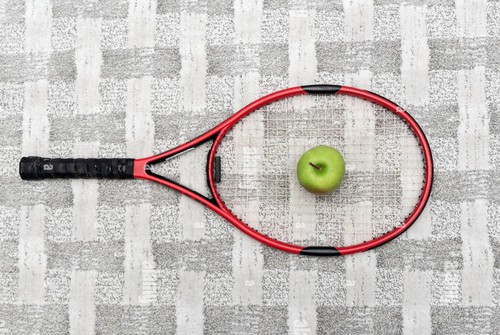} \\
        \small VLM: \hfill  apple 99 \% & \small \hfill  tennis ball 89 \%\\
        \small \strr{} fusion:  \small  \bf apple  & \\
    \end{tabular} 
    \caption{VLM (CLIP-B)
    -- zero-shot recognition with ground truth prompts and selected distractors. In the top example, recognition fails on the foreground (left, crop of a tight object bounding box). In the bottom, it fails on the full image (right).  The proposed \strr{} fusion is correct both times.
    }
    \label{fig:vlms_context}
\end{figure}

However, as Figure~\ref{fig:teaser} illustrates, context may play a critical role in object recognition~\cite{torralba2003contextual,divvala2009empirical,oliva2007role,acharya2022detecting,taesiri2024imagenet,zhu2016object,zitnick2014edge,picek2024animal}. 
Certain classes are difficult to recognize from \fg{} features alone without the supporting contextual information provided by the \bg{}. While large-scale pretraining improves robustness to some extent, recent work \cite{wang2025sober} shows that even \acp{vlm} like CLIP remain sensitive to \bg{} distribution shifts. Figure~\ref{fig:vlms_context} presents two examples: one where the context enables correct recognition, the other where misleading \bg{} causes an incorrect prediction despite a clear \fg{} object.\footnote{CLIP-B predictions are from the online demo at \url{https://huggingface.co/spaces/merve/compare_clip_siglip}.}
The nuanced role of \bg{} is overlooked in recent object recognition literature \cite{barbu2019objectnet,xiao2020noise,shetty2019not,ghosh2024aspire,wang2022clad,barbu2019objectnet,xiao2020noise,shetty2019not,ghosh2024aspire,wang2022clad}. Commonly, frequent co-occurrences of \fg{} and \bg{} are dismissed as ``spurious correlations'', a characterization we challenge as it ignores the important contribution of context to recognition.

We propose a novel approach to object recognition.  It treats localization as an integral part of the recognition process, rather than something more challenging that should only follow classification.
Our experiments show that zero-shot \fg{} localization or even segmentation as part of recognition is often feasible with modern methods \cite{kirillov2023segment,ravi2024sam,ke2024segment,zhao2023fast,radford2021learning,liu2023grounding},
particularly in the context of fine-grained recognition.
In the more general settings of datasets like ImageNet \cite{deng2009imagenet, russakovsky2015imagenet} where images may contain many different objects,
we demonstrate the potential of our approach by relying on GT prompts for object detection but without leaking the GT information into the classification model.

We first experimentally confirm that over-reliance on \bg{} significantly hurts model robustness.
We show that a straightforward approach -- zero-shot \bg{} removal --  is a strong baseline.
It outperforms or matches standard full image  (\full{}) modelling on a broad range of benchmarks. On the Spawrious \cite{lynch2023spawrious} domain generalization benchmark
it outperforms all state-of-the-art approaches that limit the influence of the \bg{} by modifying their training procedure, often relying on additional \bg{} annotations.

We proceed to show that by robustly incorporating \bg{} information in form of the standard context-aware modelling
into the \fg{}-only recognition pipeline, 
the ``Localize to Recognize Robustly" (\strr{})  method can leverage the \bg{} and further improve on in-distribution evaluation data, without loss of robustness to \bg{} distribution shifts.

We further evaluate \strr{} with non-parametric fusion on zero-shot object classification with multimodal \acp{vlm}.
The method consistently improves the performance of diverse CLIP-like models on all datasets, including the recently introduced state-of-the-art SigLIP2 \cite{tschannen2025siglip}. Notably, the performance of BioCLIP \cite{stevens2024bioclip} on the extremely challenging FungiTastic \cite{picek2024fungitastic} dataset doubles from 19 to 38 \%.

The \strr{} approach offers additional advantages.  The decomposition opens new possibilities for \bg{} modelling, such as leveraging large pretrained models with strong representations, like DINO \cite{oquab2023dinov2} and CLIP \cite{radford2021learning}, or incorporating diverse data sources, such as tabular metadata related to \bg{}. This allows the \bg{} component to capture context more effectively without extensive additional training, enhancing recognition in highly-varied environments.

The main contributions of this work are:
\begin{enumerate}
    \item Introducing \strr{}, an object classification approach that models foreground 
    (\fg{}) independently of the context-aware \full{} (which includes \bg{}), 
    enabling both robust and context-aware classification.
    \fg{} and \bg{} representations are combined through a simple, interpretable fusion module.
    \item Demonstrating that zero-shot detection (without additional training data) can now be integrated into object recognition across a wide range of finegrained datasets.
    \item  Establishing \fg{} as a strong baseline for \bg{} suppression, improving performance of supervised classifiers across all benchmarks.
    \item Showing that our approach improves on in-domain data while maintaining robustness to background shifts.
    \item The same idea applied to zero-shot classification with large-scale \acp{vlm} significantly and consistently boosts the performance across multiple benchmarks. 
\end{enumerate}

\section{Related work}
\label{sec:related}

\textbf{Complementary role of \fg{} and \bg{}.} Inspired by human vision, pioneering studies in object detection \cite{torralba2003contextual,divvala2009empirical,oliva2007role} emphasize the interdependence
between \fg{} and \bg{}. These works examine various types of contextual information and demonstrate how contextual cues provide critical insights for recognition, sometimes more so than the object itself. 
Acharya et al.~\cite{acharya2022detecting} detect out-of-context objects through context provided by other objects within a scene, modelling co-occurrence through a Graph Neural Network (GNN).

In a recent study, Taesiri et al.~\cite{taesiri2024imagenet} dissect a subset of the ImageNet dataset \cite{russakovsky2015imagenet} into \fg{}, \bg{}, and \full{} image variants using ground truth bounding boxes.
A classifier is trained on each dataset variant, finding that the \bg{} classifier successfully identifies nearly 75\% of the images misclassified by the \fg{} classifier. Additionally, they demonstrate that employing zooming as a test-time augmentation markedly improves recognition accuracy.

Closely related to our approach, Zhu et al.~\cite{zhu2016object} advocate for independent modelling of \fg{} and \bg{} with post-training fusion. Unlike our method, which leverages recent advancements in zero-shot detection, their approach requires ground truth masks. A ground-truth-free approach is also proposed, but it consists of averaging 100 edge-based bounding box proposals for each classifier \cite{zitnick2014edge}. This is not only extremely costly but also benefits heavily from ensembling, not necessarily \fg{}-\bg{} decomposition. The experiments are limited to a subset of a single dataset and weaker baselines. In contrast, our work demonstrates the relevance and effectiveness of independent \fg{} modelling fused with context-aware prediction in modern settings, even in the context of large-scale vision-language models.

Picek et al.~\cite{picek2024animal} investigate the role of \fg{} features and contextual metadata cues, such as time and location, in animal re-identification tasks. Unlike our general approach, their experiments specifically require the presence of ground-truth metadata, focus on niche applications and handcraft the \bg{} models.

Asgari et al.~\cite{asgari2022masktune} propose `MaskTune', a method which promotes the learning of a diverse set of features by masking out discriminative features identified by pre-training, without explicitly categorizing these features as \fg{} or \bg{}.

\textbf{Background suppression.} Excessive reliance on \bg{} has a detrimental impact on classifier robustness to distribution shifts  \cite{moayeri2022comprehensive,xiao2020noise,bhatt2024mitigating,barbu2019objectnet,shetty2019not}.
In response, numerous strategies have been developed to mitigate this over-reliance by suppressing \bg{} during classification.
These methods typically involve regularizing classifier training to emphasize \fg{} features, either through the use of ground-truth segmentations or attention maps \cite{bhatt2024mitigating,yang2024significant,chou2023fine,aniraj2023masking}.
This enhances \fg{} representation but prevents the classifier from learning \bg{} cues that are necessary when \fg{} is ambiguous. Moreover, when \fg{}-\bg{} correlations are strong, reliance on attention maps for segmentation proves problematic, as the attention often highlights the \bg{} \cite{moayeri2022hard}.

Another group of methods involves training classifiers on images with manipulated or out-of-distribution backgrounds to reduce \bg{} dependency \cite{barbu2019objectnet,xiao2020noise,shetty2019not,ghosh2024aspire,wang2022clad}.
This technique results in complete disregard of \bg{} information or necessitates the modelling of \fg{}-\bg{} combinations for effective training, but it is not clear how to choose the optimal \bg{} distribution.

Disentangling \fg{} and context-aware modelling eliminates the need for \bg{} suppression.

\textbf{\fg{}-\bg{} in other tasks}
In the context of image segmentation, Mask2Former \cite{cheng2022masked} also adopts the \bg{} suuppression approach implemented by masking out \bg{} tokens in cross attention with queries inside the decoder to speed up convergence. The context is still incorporated in self-attention layers. A similar camouflage \bg{} approach is adopted in \cite{luo2023camouflaged}. More recently, Cutie \cite{cheng2024putting} extends this masked attention approach by separating the semantics of
the foreground object from the background for video object segmentation, focusing half of the object queries on the \fg{} and half on the \bg{}. While \fg{} only masked attention improves over standard attention, the \fg{}-\bg{} masked attention outperforms both, showing the importance of \bg{} information

Unlike in image classification, the field of image segmentation and tracking combines \bg{} suppression with contextual information, similarly to what we propose, but none adopts the independent \fg{} and context-aware \full{} modelling approach with robust fusion.

\textbf{Reliance on \bg{} in VLMs} is analyzed by \cite{wang2025sober}. A dataset of animals, where each animal is associated with two kinds of \bg{}, a `common' one and a `rare' one, where CLIP performance on the `rare' \bg{} drops significantly.

\textbf{Zero-shot localization.} 
Recent advancements in large vision-language models \cite{radford2021learning,liu2023grounding} and class-agnostic, promptable image detection and segmentation \cite{kirillov2023segment,ravi2024sam,ke2024segment,zhao2023fast} now facilitate zero-shot object localization of a wide range of objects without knowing their (finegrained) class. This enables localization and effective \fg{}-\bg{} separation across a variety of image classification datasets.

Our methodology leverages these advances and seamlessly integrates robustness against unseen \bg{}s and utilization of the contextual information in \bg{}. 

\section{Method}
\label{sec:method}
\begin{figure}
        \includegraphics[trim={0 0cm 0 1cm},clip,width=\linewidth]
            {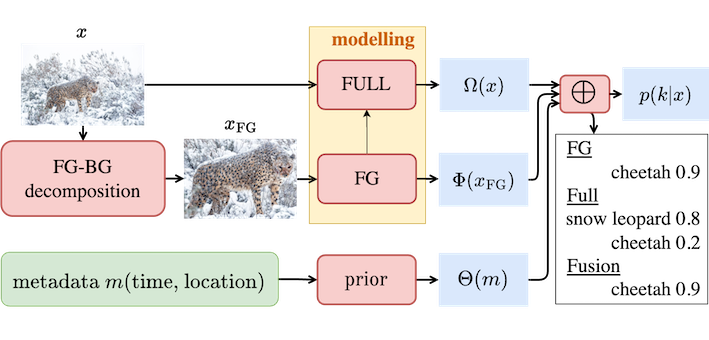}
        \caption{The ``Localize to Recognize Robustly'' approach to context-aware recognition
        -- \strr{} -- proceeds in three stages: (1) decomposition of image $x$ into \fg{} and \bg{} by zero-shot detection, possibly exploiting the predictions of \full{} for prompt generation (2) independent modelling of the \fg{} and the context-aware \full{} (original image), which also serves as a fallback option when detection fails, and (3) fusion that robustly combines the representations from stage (2) to form the output prediction $p(k|x)$.}
        \label{fig:full_pipeline}
\end{figure}

We propose a novel approach to object recognition that decouples the modelling of the \fg{} and the context-aware \full{} representation of an image and then combines them in a lightweight interpretable module. Our approach consists of three stages, see Figure \ref{fig:full_pipeline}: 1. Image decomposition, 2. \fg{} and \full{} appearance modelling, and 3. Fusion.  

\subsection{Image decomposition}
\label{sec:imdec}
The goal of this stage is to localize the pixels representing the target object $x_{\text{FG}}$. The complement is the background context, $x_{\text{BG}}$.
The decomposition relies on a zero-shot object detection model (referred to as $f_\text{D}$) such as OWL \cite{minderer2024scaling, minderer2022simple} or GroundingDINO \cite{liu2023grounding}. These models are prompted by a dataset-specific text prompt $p$. 
  
The operation of the image decomposition module can be described as 
\begin{equation} \label{x_fg_bg_def}
    x_{\text{FG}}, x_{\text{BG}} = f_\text{D}(x, p)
\end{equation}
\textbf{Detector prompts.}
For each dataset, we generate an embedding created from either a single text meta-prompt, or an average of the embeddings of multiple ones. This works well in the case of fine-grained datasets where the objects belong to a specific meta-class (e.g., recognizing dog breeds or mushroom species). The detection in such cases is easy -- a generic meta-prompt representing all classes (e.g., ``dog'' or ``mushroom'') suffices. 
\textbf{Oracle prompts:} In experiments with general multi-object (and often also multi-label) datasets like ImageNet, we do not have a generally applicable solution. To show the potential of our decomposition approach, we pre-compute masks for all the datasets based on prompting each image with the text of its GT label.

Detailed settings for each dataset, together with a broader discussion and experiments with fully automated approaches, can be found in the Supplementary.

\textbf{Fallback.} \strr{} relies on successful decomposition into \fg{} and \bg{}. Problematic detection can be flagged when the detector output is empty or the confidence falls below a threshold. In such cases, the output of \strr{} is the standard \full{} image prediction.

\subsection{Subject and context-aware modelling}
We opt for an approach where both the \fg{} and \full{} models $\Phi$ and $\Omega$, respectively, output the per-class probability $p(k|x_{\text{FG}}) = \Phi(x_{\text{FG}})$ and $p(k|x_{\text{FULL}})=\Omega(x_{\text{FULL}})$.
 
Another option explored in our experiments is the usage of a different modality representing the \bg{}, in our case tabular metadata \cite{berg2014birdsnap,Picek_2022_WACV}.

Thanks to the decoupling of the \fg{} and \full{} modelling, the \fg{} classifier can not learn \bg{} shortcuts. 
It also increases interpretability - for instance, if we encounter an object from a well-known class in an unfamiliar environment and $p(k|x_{\text{FULL}})$ is expected to be low while the probability $p(k|x_{\text{FG}})$ is expected to be much higher.

\subsection{Fusion modelling} \label{sec:fusion}
The fusion model is designed to combine the outputs of base classifiers, typically \fg{} and \full, but we also experiment with \bg{} (removing the \fg{} pixels from \full). The fusion model’s optimization is independent of the optimization of the fused models, simplifying the task. 
The fusion models are designed with interpretability in mind. 

The fusion module can combine various models (e.g., \fg{}+\bg{} or \fg{}+{\scshape full}). 
Let two pretrained models be denoted as $ \Phi_1 $ and $ \Phi_2 $, which output logit vectors $ \Phi_i(x) = z_i \in \mathbb{R}^C $. Applying softmax activations yields per-class confidences $\sigma(z_i)$. Predictions and their confidences are obtained by $ \hat{y}_i = \argmax_{k \in \{1 \dots C\}} z_i^{(k)} $ and $ \hat{p}_i = \sigma(z_i)^{(\hat{y}_i)} $.

Since deep neural networks are known to be poorly calibrated, potentially hindering model confidence comparisons, temperature-scaled logits \cite{guo2017calibration,frenkel2021network} are considered whenever applicable. Details and more fusion approaches are presented in Appendix \ref{ap:methods:comb}.

\noindent \textbf{Higher confidence} $\mathbf \oplus_{\text{max}}$: Selects the prediction with the highest confidence, setting $ \hat{p} = \max (\hat{p}_1, \hat{p}_2 ) $.

\noindent \textbf{Robust prediction} $\mathbf \oplus_{\text{R}}$: Selects $ \hat{y}_1 $ if $ \hat{p}_1 > t $, otherwise $ \max (\hat{p}_1, \hat{p}_2 ) $. The parameter $t$, where $t > 0$, can be optimized to maximize accuracy on the validation set or manually set to limit the influence of $\Phi_2$ (typically \bg{}). 

\noindent \textbf{Weighted logits} $\mathbf \oplus_{\text{WL}}$: Learns per-class weights $ w_1, w_2 \in \mathbb{R}^C $, combining logits as $ w_1 z_1 + w_2 z_2 $.
This approach trades off some of the interpretability for increased flexibility.

\section{Implementation details}
\label{sec:impl}
We provide two sets of experiments. The first one is conducted in the standard supervised training setup and the second one concerns large-scale pretrained \acp{vlm} in a zero-shot recognition setup.
Additional details concerning the datasets and models are provided in the Appendix.

\subsection{Datasets}
 We evaluate the \strr{} recognition approach on 6~classification datasets, three of which are fine-grained:
 
\noindent\textbf{FungiTastic (Fungi)} \cite{picek2024fungitastic}: A challenging fine-grained fungi species dataset with complex \fg{}-\bg{} relationships. The \bg{} can be helpful in some cases but may be constant or less informative in others.
    
\noindent\textbf{ImageNet-1K (IN-1K)} \cite{russakovsky2015imagenet}: A large dataset of diverse 1000 classes with diverse \fg{}-\bg{} relationships, many of them fine-grained. While IN-1K is the gold standard for recognition model evaluation, it is known to contain many issues~\cite{kisel2024flaws}. Therefore, we also evaluate on a `clean labels' subset which only contains images where previous works correcting the dataset agree on the label \cite{kisel2024flaws}.

\noindent \textbf{Hard ImageNet (HIN)} \cite{moayeri2022hard}: A subset of 15 IN-1K classes with strong \fg{}-\bg{} correlations. We also introduce two new test sets, Long Tail (HIN-LT) and Constant (HIN-CT), containing unusual or constant \bg{}s.

\noindent\textbf{CounterAnimal} \cite{wang2025sober}: A dataset of 45 animal classes from IN-1K with images from the iNaturalist dataset. Each image is further labelled based on the \bg{} as `common' or `rare'. 
    
\noindent\textbf{Spawrious (Spaw)} \cite{lynch2023spawrious}: A synthetic dog-breed classification dataset introduced for domain generalization. Each class is associated with a specific \bg{} type in the training set, but the \bg{} distribution changes in the test set.

\noindent\textbf{Stanford Dogs} \cite{KhoslaYaoJayadevaprakashFeiFei_FGVC2011}: A dataset where the \bg{} plays no obvious role in breed identification.

For datasets without a validation set ({Dogs} and Spaw), we reserve 10-15 \% of the training set for validation. For ImageNet-1K, we adopt the official validation set as the test set, a common practice in the literature.

\subsection{Supervised classification} \label{exp:setup}
\noindent\textbf{Evaluation.} Recognition accuracy is used as the main evaluation metric. For the highly imbalanced FungiTastic, macro-averaged accuracy (mean of per-class accuracies) is reported.  The result is an average of five models with different seeds, except for ImageNet-1K where we use a single checkpoint from Timm.

\noindent\textbf{Training - base classifiers.}
An independent classifier is learnt for \fg{} and \full{} (also \bg{} for analysis). While this  is not the most efficient approach -- doubling the cost of training and inference -- it gives us insights into how much can be learnt from different input kinds without being obfuscated by the impact of data augmentation, for example. A unified model with a shared backbone can be adopted in practice. 

All models are based on the ConvNeXt V2-Base \cite{woo2023convnext, liu2022convnet} architecture from Timm \cite{rw2019timm}, pretrained with a fully convolutional masked autoencoder (FCMAE) and fine-tuned on ImageNet-1k, unless indicated otherwise. The only exception is the ImageNet-1K dataset where we adopt the smaller ConvNeXt V2-Tiny variant for faster training.
The input size is $ 224 \times 224 $ and the batch size is $128$ for all datasets.

We train models for each of the following inputs: \full{} images, \fg{} inputs (cropped bounding box padded to a square with a constant value to prevent distorting aspect ratios), and \bg{} inputs (with \fg{} shape obtained by prompting SAM \cite{kirillov2023segment} with the bounding box masked out). Each is trained with five different seeds and the results are averaged unless stated otherwise. Experiments with additional \fg{} and \bg{} representation are provided in the Appendix.

\noindent\textbf{Fusion models.}
Fusion models combine base classifier outputs as per Section \ref{sec:fusion}. The standard fusion combines \fg{} and \bg{} (+\full{} image classifiers as fallback option), though alternative combinations (e.g., \fg{} + \full{}) and different seed variations are also tested. 

\subsection{Vision-Language Models}
We adopt the state-of-the-art SigLIP2 \cite{tschannen2025siglip} (so400m-patch14-256 variant) for main experiments, with the exception of the FungiTastic dataset, where evaluating general-purpose models is not meaningful, their performance is very low regardless of the model. Instead, we adopt the BioCLIP \cite{stevens2024bioclip} model for this dataset.

Unlike in the experiments with supervised models, no models are trained and there are no hyper-parameters; everything is zero-shot. We adopt the simplest `maximum confidence' fusion $\oplus_{\max}$ in all experiments.

\full{}, \fg{} and \bg{} are processed by the same \ac{vlm} and all results come from a model trained with the same seed since only one is publicly available. The input resolution to the models is $256 \times 256$. \fg{} inputs are padded to a square the same way as for supervised classifiers. 
Compared to standard classification,
the model processes up to twice the number of images at inference. 

\noindent\textbf{Text prompts}  For each class $c$ with a class name $n_c$, an embedding of the text `A photo of a $n_c$' is precomputed by the text encoder,
which serves as the class prototype.
Each image is then classified based on the nearest class prototype to the image embedding.
We adopt the official class names provided by the dataset authors,
no optimization of the class names was performed.

\noindent\textbf{Evaluation.} We report the macro-averaged accuracy (mean of per-class accuracies) for all datasets.

\section{Results}
\label{sec:experiments}
\subsection{Supervised training}

\begin{table*}[bth]
\small
\setlength{\tabcolsep}{4pt}
\centering
\begin{tabular}{l rrr r r rr r r}
    \toprule
    & \multicolumn{3}{c}{\textbf{HardImageNet*}} & \multicolumn{1}{c}{\textbf{Dogs}} & \multicolumn{1}{c}{\textbf{Spaw}} & \multicolumn{2}{c}{\textbf{ImageNet-1K*}} & \multicolumn{1}{c}{\textbf{Fungi}} & \multicolumn{1}{c}{\textbf{mean}} \\
        & \multicolumn{1}{c}{Original} & \multicolumn{1}{c}{Constant} & \multicolumn{1}{c}{Long-Tail} & \multicolumn{1}{c}{Original} & \multicolumn{1}{c}{Original} & \multicolumn{1}{c}{Original} & \multicolumn{1}{c}{Clean} & \multicolumn{1}{c}{Original} & \\
    \cmidrule(r){2-4} \cmidrule(r){5-5} \cmidrule(r){6-6} \cmidrule(r){7-8} \cmidrule(r){9-9} \cmidrule(r){10-10}
    \full{} &   97.33 &   90.51 &   81.33 &   90.28 &   43.20 &   82.35 &   92.01 &   43.17 &   77.52 \\
    \fg{} & {\scriptsize\textcolor{ForestGreen}{+0.46}} 97.79 & {\scriptsize\textcolor{red}{-0.41}} 90.10 & {\scriptsize\textcolor{ForestGreen}{+4.60}} 85.93 & {\scriptsize\textcolor{ForestGreen}{+0.97}} 91.25 & {\scriptsize\textcolor{ForestGreen}{+48.11}} 91.31 & {\scriptsize\textcolor{ForestGreen}{+3.21}} 85.56 & {\scriptsize\textcolor{red}{-0.02}} 91.99 & {\scriptsize\textcolor{red}{-0.08}} 43.09 & {\scriptsize\textcolor{ForestGreen}{+7.11}} 84.63 \\
    \bg{} & {\scriptsize\textcolor{ForestGreen}{+0.51}} 97.84 & {\scriptsize\textcolor{red}{-16.57}} 73.94 & {\scriptsize\textcolor{red}{-1.60}} 79.73 & {\scriptsize\textcolor{red}{-38.94}} 51.34 & {\scriptsize\textcolor{red}{-40.58}} 2.62 & {\scriptsize\textcolor{red}{-9.11}} 73.24 & {\scriptsize\textcolor{red}{-10.73}} 81.28 & {\scriptsize\textcolor{red}{-19.41}} 23.76 & {\scriptsize\textcolor{red}{-17.05}} 60.47 \\
    \midrule
    \fg{}$\oplus_{*}$\bg{} & {\scriptsize\textcolor{ForestGreen}{+1.66}} 98.99 & {\scriptsize\textcolor{ForestGreen}{+1.61}} 92.12 & {\scriptsize\textcolor{ForestGreen}{+8.67}} 90.00 & {\scriptsize\textcolor{ForestGreen}{+0.99}} 91.27 & {\scriptsize\textcolor{ForestGreen}{+22.60}} 65.80 & {\scriptsize\textcolor{ForestGreen}{+4.78}} 87.13 & {\scriptsize\textcolor{ForestGreen}{+1.29}} 93.30 & {\scriptsize\textcolor{ForestGreen}{+2.48}} 45.65 & {\scriptsize\textcolor{ForestGreen}{+5.51}} 83.03 \\
    \fg{}$\oplus_{\text{max}}$\bg{} & {\scriptsize\textcolor{ForestGreen}{+1.44}} 98.77 & {\scriptsize\textcolor{ForestGreen}{+0.60}} 91.11 & {\scriptsize\textcolor{ForestGreen}{+8.67}} 90.00 & {\scriptsize\textcolor{red}{-6.50}} 83.78 & {\scriptsize\textcolor{red}{-17.30}} 25.90 & {\scriptsize\textcolor{ForestGreen}{+4.04}} 86.39 & {\scriptsize\textcolor{ForestGreen}{+1.01}} 93.02 & {\scriptsize\textcolor{red}{-1.62}} 41.55 & {\scriptsize\textcolor{red}{-1.21}} 76.31 \\
    \fg{}$\oplus_{\text{WL}}$\bg{} & {\scriptsize\textcolor{ForestGreen}{+1.60}} 98.93 & {\scriptsize\textcolor{gray}{0.00}}  90.51 & {\scriptsize\textcolor{ForestGreen}{+9.11}} 90.44 & {\scriptsize\textcolor{red}{-3.07}} 87.21 & {\scriptsize\textcolor{red}{-15.49}} 27.71 & {\scriptsize\textcolor{ForestGreen}{+4.78}} 87.13 & {\scriptsize\textcolor{ForestGreen}{+1.29}} 93.30 & {\scriptsize\textcolor{ForestGreen}{+2.48}} 45.65 & {\scriptsize\textcolor{ForestGreen}{+0.09}} 77.61 \\
    \fg{}$\oplus_{\text{R}}$\bg{} & {\scriptsize\textcolor{ForestGreen}{+0.78}} 98.11 & {\scriptsize\textcolor{ForestGreen}{+0.40}} 90.91 & {\scriptsize\textcolor{ForestGreen}{+5.66}} 86.99 & {\scriptsize\textcolor{ForestGreen}{+0.66}} 90.94 & {\scriptsize\textcolor{ForestGreen}{+48.05}} 91.25 & {\scriptsize\textcolor{ForestGreen}{+4.22}} 86.57 & {\scriptsize\textcolor{ForestGreen}{+1.17}} 93.18 & {\scriptsize\textcolor{red}{-1.31}} 41.86 & {\scriptsize\textcolor{ForestGreen}{+7.45}} 84.98 \\
    \midrule
    \fg{}$\oplus_{\text{*}}$\full{} & {\scriptsize\textcolor{ForestGreen}{+1.52}} 98.85 & {\scriptsize\textcolor{red}{-0.41}} 90.10 & {\scriptsize\textcolor{ForestGreen}{+7.43}} 88.76 & {\scriptsize\textcolor{ForestGreen}{+1.88}} 92.16 & {\scriptsize\textcolor{red}{-0.20}} 43.00 & {\scriptsize\textcolor{ForestGreen}{+4.69}} 87.04 & {\scriptsize\textcolor{ForestGreen}{+1.75}} 93.76 & {\scriptsize\textcolor{ForestGreen}{+5.10}} 48.27 & {\scriptsize\textcolor{ForestGreen}{+2.72}} 80.24 \\
    \fg{}$\oplus_{\text{max}}$\full{} & {\scriptsize\textcolor{ForestGreen}{+1.39}} 98.72 & {\scriptsize\textcolor{red}{-0.41}} 90.10 & {\scriptsize\textcolor{ForestGreen}{+7.79}} 89.12 & {\scriptsize\textcolor{ForestGreen}{+1.41}} 91.69 & {\scriptsize\textcolor{ForestGreen}{+33.46}} 76.66 & {\scriptsize\textcolor{ForestGreen}{+3.87}} 86.22 & {\scriptsize\textcolor{ForestGreen}{+1.49}} 93.50 & {\scriptsize\textcolor{ForestGreen}{+2.00}} 45.17 & {\scriptsize\textcolor{ForestGreen}{+6.38}} 83.90 \\
     \fg{}$\oplus_{\text{WL}}$\full{} & {\scriptsize\textcolor{ForestGreen}{+1.68}} 99.01 & {\scriptsize\textcolor{ForestGreen}{+0.20}} 90.71 & {\scriptsize\textcolor{ForestGreen}{+8.05}} 89.38 & {\scriptsize\textcolor{ForestGreen}{+1.78}} 92.06 & {\scriptsize\textcolor{ForestGreen}{+33.75}} 76.95 & {\scriptsize\textcolor{ForestGreen}{+4.69}} 87.04 & {\scriptsize\textcolor{ForestGreen}{+1.75}} 93.76 & {\scriptsize\textcolor{ForestGreen}{+5.10}} 48.27 & {\scriptsize\textcolor{ForestGreen}{+7.12}} 84.65 \\
    \fg{}$\oplus_{\text{R}}$\full{} & {\scriptsize\textcolor{ForestGreen}{+0.78}} 98.11 & {\scriptsize\textcolor{red}{-0.61}} 89.90 & {\scriptsize\textcolor{ForestGreen}{+5.84}} 87.17 & {\scriptsize\textcolor{ForestGreen}{+1.38}} 91.66 & {\scriptsize\textcolor{ForestGreen}{+48.06}} 91.26 & {\scriptsize\textcolor{ForestGreen}{+4.03}} 86.38 & {\scriptsize\textcolor{ForestGreen}{+1.47}} 93.48 & {\scriptsize\textcolor{ForestGreen}{+1.56}} 44.73 & {\scriptsize\textcolor{ForestGreen}{+7.81}} 85.34 \\
        \bottomrule
    \end{tabular}

\caption{
 Recognition accuracy of \fg{}, \bg{}, \full{} and of several fusion models on Hard ImageNet, Stanford Dogs and Spawrious. For FungiTastic, which is highly imbalanced, the mean class accuracy is reported.
 The $\oplus_{\text{*}}$ fusion method is selected on the validation set. *Results with oracle detection obtained by GT prompting}
\label{tab:summary}
\end{table*}

\begin{figure}[bt]
    \centering
        \setlength{\tabcolsep}{1pt}
    \begin{tabular}{cc}
         \includegraphics[width=0.49\linewidth,]{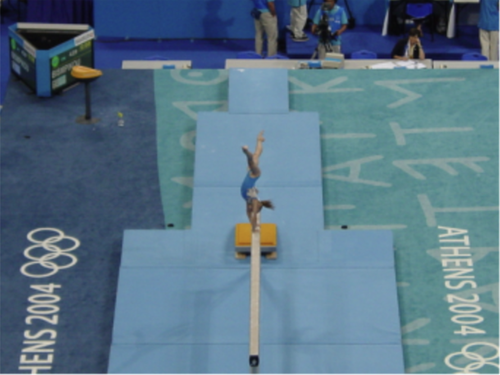} & \includegraphics[width=0.49\linewidth,]{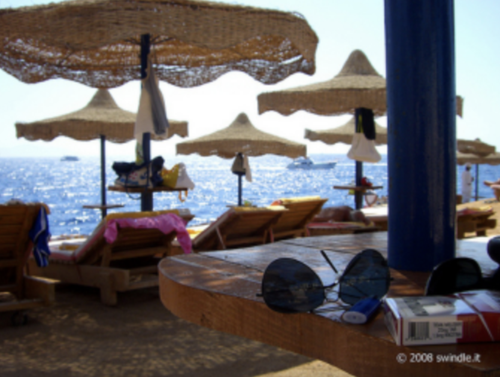} \\
        \colorbox{ForestGreen!30}{balance beam} / \colorbox{red!30}{horiz. bar}
        &
        \colorbox{ForestGreen!30}{sunglasses} / \colorbox{red!30}{patio} \\
    \end{tabular}

    \caption{The unexpected role of shape in \bg{} modelling. When investigating the results on the Hard ImageNet dataset, many examples where found where  full image prediction is {\setlength{\fboxsep}{1pt}\colorbox{red!30}{incorrect}} but \textit{both} \fg{} and \bg{} (with shape) predictions are {\setlength{\fboxsep}{1pt}\colorbox{ForestGreen!30}{correct}} 
    Possible explanation:  the mask provides the \bg{} model with information about the location of the target object and 
    its shape,  information not available to unlike the full image model.
    }
    \label{fig:hin_bg_shape}
\end{figure}

\begin{figure}[bt]
    \centering
        \setlength{\tabcolsep}{0.5pt}
    \begin{tabular}{cc}
         \includegraphics[width=0.45\linewidth,]{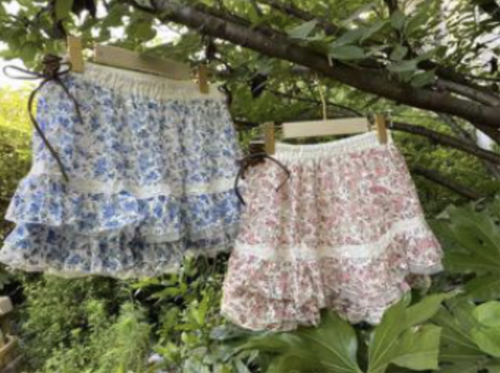} & \includegraphics[width=0.48\linewidth,]{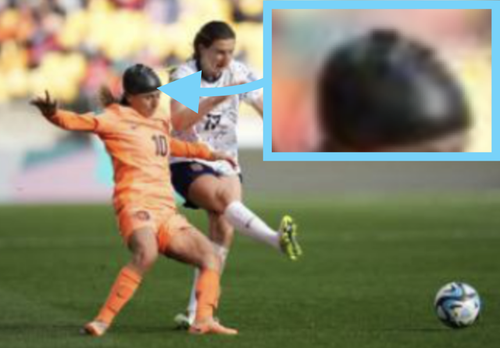} 
         \\
        \colorbox{ForestGreen!30}{\small{miniskirt}} / \colorbox{red!30}{\small{howler monkey}}
        &
        \colorbox{ForestGreen!30}{\small{swim. cap}} / \colorbox{red!30}{\small{baseball player}} \\
    \end{tabular}

    \caption{Examples where \fg {} model is {\setlength{\fboxsep}{1pt}\colorbox{ForestGreen!30}{correct}} and both full image and \bg{} models are {\setlength{\fboxsep}{1pt}\colorbox{red!30}{incorrect}} on Hard ImageNet - Long Tail.}
    \label{fig:hin_fg_imp}
\end{figure}

\begin{table}[tbh]

\setlength{\tabcolsep}{4pt}

\begin{tabular}{lrlr}
    \toprule
    
    ERM \cite{vapnik1991principles} & {\scriptsize\textcolor{ForestGreen}{+6.14}} 77.49  & JTT \cite{liu2021just} & {\scriptsize\textcolor{ForestGreen}{+18.89}} 90.24  \\
    GroupDRO \cite{sagawa2019distributionally} & {\scriptsize\textcolor{ForestGreen}{+9.23}} 80.58  & Mixup \cite{xu2020adversarial} & {\scriptsize\textcolor{ForestGreen}{+17.13}} 88.48  \\
    IRM [C] & {\scriptsize\textcolor{ForestGreen}{+4.10}} 75.45  & Mixup \cite{yao2022improving} & {\scriptsize\textcolor{ForestGreen}{+17.29}} 88.64  \\
    CORAL \cite{sun2016deep} & {\scriptsize\textcolor{ForestGreen}{+18.31}} 89.66 & \multicolumn{2}{c}{\textbf{\strr{} } (our)} \\
    \cmidrule(l){3-4}
    CausIRL \cite{chevalley2022invariant} & {\scriptsize\textcolor{ForestGreen}{+17.97}} 89.32  & \full{} & 
         71.35  \\
    MMD-AAE \cite{li2018domain} & {\scriptsize\textcolor{ForestGreen}{+7.46}} 78.81  & $\text{\fg{}}_{\text{C}}$ & {\scriptsize\textcolor{ForestGreen}{\underline{+23.65}}} \underline{95.00} \\
    Fish \cite{shi2021gradient} & {\scriptsize\textcolor{ForestGreen}{+6.16}} 77.51  & $\text{\fg{}}_{\text{M}}$ & {\scriptsize\textcolor{ForestGreen}{\textbf{+24.24}}} \textbf{95.59} \\
    VREX \cite{krueger2021out} & {\scriptsize\textcolor{ForestGreen}{+13.34}} 84.69  & \bg{} & {\scriptsize\textcolor{red}{-62.45}}\hspace{1ex} 8.90  \\
    W2D \cite{huang2022two} & {\scriptsize\textcolor{ForestGreen}{+10.59}} 81.94  & \fg{}$_{\text{C}} \oplus_{\text{R}}$ \bg{} & {\scriptsize\textcolor{ForestGreen}{+15.43}} 86.78  \\
    \bottomrule
\end{tabular}
    \caption{Spawrious \cite{lynch2023spawrious}, a dataset with an adversarial \bg{} shift -- comparison to domain generalization methods.
    The \textbf{best} and \underline{second best} results are highlighted. $\text{FG}_{\text{C}}$ denotes cropping based on segmentation bbox, $\text{FG}_{\text{M}}$ also removes the \bg{} pixels from $\text{FG}_{\text{C}}$.
    All methods are initialized with the same ResNet50 model. 
    }
    \label{tab:spaw:r50}
\end{table}

An overview of both base and fusion models' results on all test datasets is provided in Table \ref{tab:summary}.

\textbf{Base models.} The standard \textbf{\full{} classification} provides a strong baseline across most datasets, with a moderate drop in performance on HIN-LT (-$16\%$) and HIN-CT (-$7\%$) compared to the original in-distribution test set. A more significant performance drop (from $99.9\%$ to $43.2\%$) is observed on Spawrious between the validation and test set, where the model overfits to \bg{}, which changes substantially between training and test sets.
The\textbf{ \fg{} model} outperforms \full{} by 7.11 \% on average, either improving or maintaining performance around the \full{} baseline on all datasets. 
As expected, the \textbf{\bg{} model} generally performs worse than \full{} and \fg{}, but still  very high due to the inclusion of the shape information. A notable exception is the Original HIN test set, where the \bg{} is so correlated with the \fg{} that the performance of \bg{} matches \full{}.

Images where \bg{} works well while \fg{} and \full{} does not are presented in \ref{fig:hin_bg_shape}, images where only \fg{} is correct are shown in \ref{fig:hin_fg_imp}.

\textbf{Fusion models.} The performance of different kinds of fusion is dataset-dependent, there is still a natural trade-off between in-domain gains and robustness to domain shift. Our findings can be summarized as: 1. The selection of the fusion method on the validation set provides close to optimal results with the exception of Spaw, where the training data does not contain any examples where the \fg{} would be needed - there is no data to train the fusion models. 2. Even non-parametric $\oplus_{\text{max}}$ fusion works well on most datasets, but it can lead to significant performance drops on others, which hints at \bg{} over-confidence. 3. Learnt fusion models like $\oplus_{\text{WL}}$ lead to the strongest results, provided enough diverse training data. 4. For maximum robustness, $\oplus_{\text{R}}$ is the best choice, as shown by its strong performance of Spaw, but the gains from context incorporation may be limited compared to the other methods.
We also explored the impact of swapping the context-aware \full{} model for the \bg{} model in fusion. It sometimes works better, likely due to the explicit shape information or stronger learnt \bg{} priors, but on most datasets, \full{} leads to bigger performance gains and is more computationally efficient, as the \bg{} model requires a segmenter.

Even though `oracle prompt' detection was used for the ImageNet experiments, the results highlight how much progress on the dataset is blocked by a) localization capabilities of the classifier (and the images being multilabel) and b) lack of robust context handling. On the original validation set, \fg{} improves over \full{} by 3.21\% and \strr{} further improves over \fg{} by ~1.5\%, reaching an accuracy of 87.04\%. The performance of ConvNext V2-B from Timm, a model 3 $\times$ larger than the Tiny variant presented here, is 84.9\% on the original test set using the center crop data augmentation, which slightly inflates the performance. 

Detailed results of the base classifiers (different \fg{}, \bg{}, and \full{} image models), as well as additional fusion models, are presented in Appendix \ref{sec:ap:exps}.

\noindent\textbf{Comparison to ensembling.}
A comparison of the \strr{} \fg{}$\oplus_{*}$\full{} approach to \full{}$\oplus_{*}$\full{} and \fg{}$\oplus_{*}$\fg{} is presented in  Table \ref{tab:fullx2_fgx2_comparisons}.  
\begin{table}
\small
\setlength{\tabcolsep}{4pt}
\begin{tabular}{l rrr r r r}
        \toprule
        & \multicolumn{3}{c}{\textbf{HardImageNet}} & \multicolumn{1}{c}{\textbf{Dogs}} & \multicolumn{1}{c}{\textbf{Spaw}} & \multicolumn{1}{c}{\textbf{Fungi}} \\
        & \multicolumn{1}{c}{O} & \multicolumn{1}{c}{CT} & \multicolumn{1}{c}{LT} & \multicolumn{1}{c}{O} & \multicolumn{1}{c}{O} & \multicolumn{1}{c}{O} \\
        \cmidrule(r){2-4} \cmidrule(r){5-5} \cmidrule(r){6-6} \cmidrule(r){7-7}
        \full{} & $97.33$ & $90.51$ & $81.33$ & $90.28$ & $43.20$ & $43.17$ \\
        \fg{}$ $ & $97.79$ & $90.10$ & $85.93$ & $91.25$ & $91.31$ & $43.09$ \\
        \midrule
        \full{}$\oplus_{*}$\full{} & 97.51 & 90.91 & 82.89  & 91.13 & 40.00
        & \textbf{48.54}\\
        \fg{}$\oplus_{*}$\fg{} & 97.91 & 89.9  & 86.58 & 91.9 & \textbf{94.62} & 47.24 \\
        \midrule
        \fg{}$\oplus_{*}$\bg{}& \textbf{98.99} & \textbf{92.12} & \textbf{90.0} & $91.27$ & 65.81 & $45.65$ \\
        \fg{}$\oplus_{*}$\full{} & $98.85$ & $90.10$ & $88.76$ & \textbf{92.16} & 43-77 & $48.27$ \\
        \bottomrule
    \end{tabular}
    \caption{Mean accuracy of \strr{} models compared to different input models ensembles. O stands for the original test set.}
    \label{tab:fullx2_fgx2_comparisons}
\end{table}

\noindent\textbf{Comparison to domain generalization methods.} To provide a fair comparison of the \strr{} approach for \bg{} influence suppression to previous domain generalization methods, we provide results of Resnet50 classifiers and compare to the results from Spawrious \cite{lynch2023spawrious} in
Table \ref{tab:spaw:r50}.
The \fg{} (\fg{}$_{\text{C}}$) model beats all the domain generalization methods by a large margin (4.8 \%). Masking out the \bg{} pixels based on a segmentation mask (\fg{}$_{\text{C}}$) improves the performance further.
The \strr{} fusion \fg{} $\oplus_{}\text{R}$ slightly reduces the performance compared to \fg{} but it still outperforms \full{} by 15\% and leaves 7 out of the 12 domain generalization methods behind.

\noindent\textbf{BG model with FungiTastic metadata}
This experiment explores an alternative approach to \bg{} modelling based on tabular metadata.
The FungiTastic dataset comes with such additional data, some of which are highly related to the \bg{} appearance. Inspired by the metadata prior model of \cite{Picek_2022_WACV,berg2014birdsnap}, we study the performance of incorporating various \bg{}-related metadata, namely the habitat, substrate and month, with the \full{} (as done by \cite{Picek_2022_WACV,berg2014birdsnap}) and \fg{} models. In summary, the method precomputes a prior probability of each class - metadata value combination and reweights the classifier predictions based on the metadata. The model assumes the appearance of the image is independent of the metadata, which is not true when the image \bg{} is included (such as in the case of \full{}). Combining with \fg{} makes the method more principled.

Results in Table \ref{tab:fungi_meta} show that all metadata kinds improve the performance of both models. The habitat helps the most, adding 3.8 \% to the 43.5 \% baseline of \full{} and 4.2 \% to the 44 \% baseline of \fg{}. For habitat and month, the improvements from metadata fusion are greater for the \fg{} than for the \full{}, even though the \fg{} already performs better than \full{}. We hypothesize this can be due to the suppression of \bg{} influence in \fg{}$_\text{C}$, leading to better \fg{}-\bg{} decoupling, as assumed by the metadata model.

\begin{table}[tbh]
\setlength{\tabcolsep}{4pt}
    \centering
    \begin{tabular}{lrrrr}
    \toprule
     & \multicolumn{1}{c}{\textbf{img}} & \multicolumn{1}{c}{\textbf{+habitat}} & \multicolumn{1}{c}{\textbf{+substrate}} & \multicolumn{1}{c}{\textbf{+month}} \\
    \midrule
    \full{} & 43.50 & 47.26 {\color{darkgreen}\scriptsize +3.77} & 45.42 {\color{darkgreen}\scriptsize +1.92} & 45.19 {\color{darkgreen}\scriptsize +1.70} \\
    \fg{} & 44.00 & 48.22 {\color{darkgreen}\scriptsize +4.22} & 45.77 {\color{darkgreen}\scriptsize +1.77} & 45.80 {\color{darkgreen}\scriptsize +1.81} \\
    \bottomrule
    \end{tabular}
    \caption{Mean class accuracy of fusion models with \bg{} representation \cite{berg2014birdsnap,Picek_2022_WACV} based on tabular metadata (habitat, substrate, month) on the FungiTatsic dataset. The increment over image-only performance is also reported. The results are averaged across 5 runs with different random seeds.}
    \label{tab:fungi_meta}
\end{table}

\subsection{Zero-shot recognition with VLMs}
\begin{table*}[bth]
\small
\centering
\setlength{\tabcolsep}{3pt}
\begin{tabular}{l rrr r r rr r r}
\toprule
 & \multicolumn{3}{c}{\textbf{HardImageNet*}} & \multicolumn{1}{c}{\textbf{Dogs}} & \multicolumn{1}{c}{\textbf{Spaw}} & \multicolumn{2}{c}{\textbf{CounterAnimal}} &  & \multicolumn{1}{c}{\textbf{Fungi}} \\
  & \multicolumn{1}{c}{Original} & \multicolumn{1}{c}{Constant} & \multicolumn{1}{c}{Long-Tail} & \multicolumn{1}{c}{Original} & \multicolumn{1}{c}{Original} & \multicolumn{1}{c}{Common} & \multicolumn{1}{c}{Rare} &
    \multicolumn{1}{c}{mean} 
  & \multicolumn{1}{c}{Original} \\
\cmidrule(r){2-4} \cmidrule(r){5-5} \cmidrule(r){6-6} \cmidrule(r){7-8} \cmidrule(r){9-9} \cmidrule(r){10-10}

\multicolumn{1}{l}{\textbf{SigLIP2-SO}} & \multicolumn{7}{r}{
\leftarrowfill}  &
    & \multicolumn{1}{r}{\textbf{BioCLIP}} \\
\cmidrule(r){1-9} \cmidrule(r){10-10}

\full{} & 95.33 & 100.00 & 96.22 & 84.11 & 95.34 & 95.50 & 89.36 & 93.69 & 18.62 \\

\bg{} & {\scriptsize\textcolor{red}{-1.86}} 93.47 & {\scriptsize\textcolor{red}{-10.54}} 89.46 & {\scriptsize\textcolor{red}{-11.12}} 85.10 & {\scriptsize\textcolor{red}{-62.37}} 21.74 & {\scriptsize\textcolor{red}{-16.92}} 78.42 & {\scriptsize\textcolor{red}{-11.49}} 84.01 & {\scriptsize\textcolor{red}{-13.47}} 75.89 & {\scriptsize\textcolor{red}{-18.25}} 75.44 & {\scriptsize\textcolor{red}{-16.43}} 2.19 \\

\fg{} & {\scriptsize\textcolor{red}{-2.53}} 92.80 & {\scriptsize\textcolor{ForestGreen}{+0.00}} 100.00 & {\scriptsize\textcolor{red}{-1.67}} 94.55 & {\scriptsize\textcolor{ForestGreen}{+0.37}} 84.48 & {\scriptsize\textcolor{ForestGreen}{+1.33}} 96.67 & {\scriptsize\textcolor{red}{-1.31}} 94.19 & {\scriptsize\textcolor{red}{-0.92}} 88.44 & {\scriptsize\textcolor{red}{-0.68}} 93.02 & {\scriptsize\textcolor{red}{-1.75}} 16.87 \\

{\sc CenterCrop} & \textcolor{red}{\scriptsize -4.00} 91.33 & \textcolor{red}{\scriptsize -12.16} 87.84 & \textcolor{red}{\scriptsize -7.03} 89.19 & \textcolor{red}{\scriptsize -3.22} 80.89 & \textcolor{ForestGreen}{\scriptsize 0.00} 95.34 & \textcolor{ForestGreen}{\scriptsize 0.17} 95.67 & \textcolor{red}{\scriptsize -1.20} 88.16 & \textcolor{red}{\scriptsize -3.92} 89.77 & \textcolor{red}{\scriptsize -0.19} 18.43 \\

\midrule
\fg{} $\oplus_{\max}$ \bg{} & {\scriptsize\textcolor{ForestGreen}{+1.34}} 96.67 & {\scriptsize\textcolor{ForestGreen}{+0.00}} 100.00 & {\scriptsize\textcolor{ForestGreen}{+1.57}} 97.79 & {\scriptsize\textcolor{ForestGreen}{+0.58}} 84.69 & {\scriptsize\textcolor{red}{-0.41}} 94.93 & {\scriptsize\textcolor{red}{-0.63}} 94.87 & {\scriptsize\textcolor{red}{-2.61}} 86.75 & {\scriptsize\textcolor{red}{-0.02}} 93.67 & {\scriptsize\textcolor{ForestGreen}{+14.92}} 33.54 \\
\fg{} $\oplus_{\max}$ \full{} & {\scriptsize\textcolor{ForestGreen}{+1.07}} 96.40 & {\scriptsize\textcolor{ForestGreen}{+0.00}} 100.00 & {\scriptsize\textcolor{ForestGreen}{+2.01}} 98.23 & {\scriptsize\textcolor{ForestGreen}{+0.97}} 85.08 & {\scriptsize\textcolor{ForestGreen}{+1.09}} 96.43 & {\scriptsize\textcolor{ForestGreen}{+0.17}} 95.67 & {\scriptsize\textcolor{red}{-1.16}} 88.20 & {\scriptsize\textcolor{ForestGreen}{+0.59}} 94.29 & {\scriptsize\textcolor{ForestGreen}{+19.18}} 37.80 \\

\bottomrule
\end{tabular}
\caption{Mean accuracy (\%) of maximum confidence fusion of \fg{} + \bg{} and \fg{} + \full{} on zero-shot classification with VLMs on different dataset test sets. Different kinds of inputs - \full{}, \fg{} and \bg{}, are also reported. BioCLIP results are reported for fungi because general-purpose VLMs perform very poorly on such complex, niche datasets. *Results with oracle detection obtained by GT prompting.}
\label{tab:vlm_comp}
\end{table*}

An overview of  zero-shot \strr{} with BioCLIP (FungiTastic) and SigLIP2 (all other datasets) results is provided in Table \ref{tab:vlm_comp}.
\strr{} improves the performance of SigLIP2 on all datasets, except for the 'rare' test set of the CounterAnimal dataset. On average, the improvement is by 0.6\% compared to \full{} from  93.69 to 94.29\%. The biggest gain is achieved on the Hard ImageNet - Long Tail test set, from 96.22 to 98.23\% (2.01\%).

The combination of \fg{} with \full{} overall outperforms \fg{} with \bg{}, possibly becuase the \bg{} inputs may be out-of-distribution for the models, or because \full{} also allows to benefit from ensembling different views of \fg{} (\fg{} + \full{} can be viewed as ($2 \times$ \fg{}) + \bg{}).
On average there is no benefit from using \fg{} only compared to \full{}, the \acp{vlm} models are more robust to \bg{} distribution shift than their supervised counterparts.

We also compare include the models performance with CenterCrop data augmentation. The results are comparable to \fg{} on CounterAnimal, a dataset with a strong center bias, but performs much worse on the other datasets, confirming the necessity of explicit localization step.

Experiments with CLIP-B and CLIP-L can be found in Appendix \ref{sec:ap:exps}, as well as more insights on the somewhat counter-intuitive negative results on the CounterAnimal datasets, where one would expect \bg{} removal to improve a lot on the `rare' subset. Part of the problem can be attributed to some classes being hard to detect well, such as thin spiders, but there are also dataset construction issues as well which obfuscate the results.

\subsection{Comparison to prior work}
In all experiments, we aim to provide a fair comparison (such as the same training and inference procedure or the same amount of hyper-parameter tuning) between all models to show the benefits of the \strr{} approach compared to an equivalent \full{} object classification model. 
We abstain from claiming state-of-the-art on any of the datasets since we beat some previous methods simply through better hyper-parameter tuning. On others, such as the Stanford Dogs dataset, our models underperform because we reserve part of the training data for validation.

\section{Conclusion}
\label{sec:conclusion}
This paper introduced ``Localize to Recognize Robustly'', \strr{}, an approach to object recognition where the benefits of context-aware recognition are combined with robustness to long-tail and out-of-domain \bg{}s.
\strr{} incorporates zero-shot object localization into the recognition process, enabling the decoupling of \fg{} and context-aware \full{} modelling. 

Our experiments demonstrate that zero-shot \bg{} removal alone is a strong baseline for supervised models across diverse datasets, consistently outperforming standard full-image models in scenarios both with and without distribution shift. Notably, on the Spawrious \cite{lynch2023spawrious} domain generalization benchmark, this approach surpassed all domain generalization baselines by a large margin -- \strr{} achieved an accuracy of 94.39\%, while the runner-up achieved 90.24\%. 

Experiments with combined modelling further show that robustly incorporating \bg{} information in form of context-aware \full{} prediction to the aforementioned baseline further improves performance on all in-domain datasets with only a small trade-off in terms of robustness to \bg{} distribution shift.

Finally, we show that the \strr{} approach with parameter-free fusion applied to \acp{vlm} improves the performance of diverse CLIP-like models, including the state-of-the-art SigLIP2. Notably, the performance of the BioCLIP model on the FungiTastic dataset doubles, highlighting the potential of this approach in the biological domain.

\noindent\textbf{Limitations.}
A primary limitation of this approach is its reliance on vision-language models in zero-shot object detectors, which may not generalize as well to very niche domains.
We also discovered that current zero-shot object detectors do not allow us to apply the methodology to a fully general setup of datasets like ImageNet where images may contain many different objects.

We focused on demonstrating the benefits of the proposed approach, but the approach adopted in experiments with supervised learning introduces additional computational complexity by requiring two classifiers, which increases overhead compared to standard classification pipelines. Nonetheless our experiments with \acp{vlm} confirm the method works even when a single model is used for different kinds of inputs.

\noindent\textbf{Future work.}
\label{par:future}
The research opens up the space for several directions of future work.
First, we envision \strr{} applied to more general settings in the context of object detection, either improving the detection classification head or building on top of class-agnostic detectors.
Another area consists of exploring other possibilities of \fg{}, \bg{} and fusion modelling.
For instance, occlusion can be removed in the \fg{} space as part of the \bg{} removal and occlusion data augmentation
 of the \fg{} input consists of simply masking out portions of the image, without needing to model different textures. 
Efficiency improvements could leverage strong pretrained representations, such as those in DINOv2 \cite{oquab2023dinov2}, to reduce computational demands. The increased computational cost can also be mitigated by only running the \full{} classifier when \fg{} is not confident.

\newpage

{
    \small
    \bibliographystyle{ieeenat_fullname}
    \bibliography{main}

\begin{thebibliography}{65}
\providecommand{\natexlab}[1]{#1}
\providecommand{\url}[1]{\texttt{#1}}
\expandafter\ifx\csname urlstyle\endcsname\relax
  \providecommand{\doi}[1]{doi: #1}\else
  \providecommand{\doi}{doi: \begingroup \urlstyle{rm}\Url}\fi

\bibitem[Acharya et~al.(2022)Acharya, Roy, Koneripalli, Jha, Kanan, and Divakaran]{acharya2022detecting}
Manoj Acharya, Anirban Roy, Kaushik Koneripalli, Susmit Jha, Christopher Kanan, and Ajay Divakaran.
\newblock Detecting out-of-context objects using contextual cues.
\newblock \emph{arXiv preprint arXiv:2202.05930}, 2022.

\bibitem[Aniraj et~al.(2023)Aniraj, Dantas, Ienco, and Marcos]{aniraj2023masking}
Ananthu Aniraj, Cassio~F Dantas, Dino Ienco, and Diego Marcos.
\newblock Masking strategies for background bias removal in computer vision models.
\newblock In \emph{Proceedings of the IEEE/CVF International Conference on Computer Vision}, pages 4397--4405, 2023.

\bibitem[Asgari et~al.(2022)Asgari, Khani, Khani, Gholami, Tran, Mahdavi~Amiri, and Hamarneh]{asgari2022masktune}
Saeid Asgari, Aliasghar Khani, Fereshte Khani, Ali Gholami, Linh Tran, Ali Mahdavi~Amiri, and Ghassan Hamarneh.
\newblock Masktune: Mitigating spurious correlations by forcing to explore.
\newblock \emph{Advances in Neural Information Processing Systems}, 35:\penalty0 23284--23296, 2022.

\bibitem[Barbu et~al.(2019)Barbu, Mayo, Alverio, Luo, Wang, Gutfreund, Tenenbaum, and Katz]{barbu2019objectnet}
Andrei Barbu, David Mayo, Julian Alverio, William Luo, Christopher Wang, Dan Gutfreund, Josh Tenenbaum, and Boris Katz.
\newblock Objectnet: A large-scale bias-controlled dataset for pushing the limits of object recognition models.
\newblock \emph{Advances in neural information processing systems}, 32, 2019.

\bibitem[Berg et~al.(2014)Berg, Liu, Woo~Lee, Alexander, Jacobs, and Belhumeur]{berg2014birdsnap}
Thomas Berg, Jiongxin Liu, Seung Woo~Lee, Michelle~L Alexander, David~W Jacobs, and Peter~N Belhumeur.
\newblock Birdsnap: Large-scale fine-grained visual categorization of birds.
\newblock In \emph{Proceedings of the IEEE conference on computer vision and pattern recognition}, pages 2011--2018, 2014.

\bibitem[Beyer et~al.(2020)Beyer, H{\'{e}}naff, Kolesnikov, Zhai, Van Den~Oord, Brain, and London]{Beyer2020AreImageNet}
Lucas Beyer, Olivier~J H{\'{e}}naff, Alexander Kolesnikov, Xiaohua Zhai, Aäron Van Den~Oord, Google Brain, and Deepmind~( London.
\newblock {Are we done with ImageNet?}
\newblock 2020.

\bibitem[Bhatt et~al.(2024)Bhatt, Das, Sigal, and N~Balasubramanian]{bhatt2024mitigating}
Gaurav Bhatt, Deepayan Das, Leonid Sigal, and Vineeth N~Balasubramanian.
\newblock Mitigating the effect of incidental correlations on part-based learning.
\newblock \emph{Advances in Neural Information Processing Systems}, 36, 2024.

\bibitem[Cheng et~al.(2022)Cheng, Misra, Schwing, Kirillov, and Girdhar]{cheng2022masked}
Bowen Cheng, Ishan Misra, Alexander~G Schwing, Alexander Kirillov, and Rohit Girdhar.
\newblock Masked-attention mask transformer for universal image segmentation.
\newblock In \emph{Proceedings of the IEEE/CVF conference on computer vision and pattern recognition}, pages 1290--1299, 2022.

\bibitem[Cheng et~al.(2024)Cheng, Oh, Price, Lee, and Schwing]{cheng2024putting}
Ho~Kei Cheng, Seoung~Wug Oh, Brian Price, Joon-Young Lee, and Alexander Schwing.
\newblock Putting the object back into video object segmentation.
\newblock In \emph{Proceedings of the IEEE/CVF Conference on Computer Vision and Pattern Recognition}, pages 3151--3161, 2024.

\bibitem[Chevalley et~al.(2022)Chevalley, Bunne, Krause, and Bauer]{chevalley2022invariant}
Mathieu Chevalley, Charlotte Bunne, Andreas Krause, and Stefan Bauer.
\newblock Invariant causal mechanisms through distribution matching.
\newblock \emph{arXiv preprint arXiv:2206.11646}, 2022.

\bibitem[Chou et~al.(2023)Chou, Kao, and Lin]{chou2023fine}
Po-Yung Chou, Yu-Yung Kao, and Cheng-Hung Lin.
\newblock Fine-grained visual classification with high-temperature refinement and background suppression.
\newblock \emph{arXiv preprint arXiv:2303.06442}, 2023.

\bibitem[Deng et~al.(2009)Deng, Dong, Socher, Li, Li, and Fei-Fei]{deng2009imagenet}
Jia Deng, Wei Dong, Richard Socher, Li-Jia Li, Kai Li, and Li Fei-Fei.
\newblock Imagenet: A large-scale hierarchical image database.
\newblock In \emph{2009 IEEE conference on computer vision and pattern recognition}, pages 248--255. Ieee, 2009.

\bibitem[Divvala et~al.(2009)Divvala, Hoiem, Hays, Efros, and Hebert]{divvala2009empirical}
Santosh~K Divvala, Derek Hoiem, James~H Hays, Alexei~A Efros, and Martial Hebert.
\newblock An empirical study of context in object detection.
\newblock In \emph{2009 IEEE Conference on computer vision and Pattern Recognition}, pages 1271--1278. IEEE, 2009.

\bibitem[Frenkel and Goldberger(2021)]{frenkel2021network}
Lior Frenkel and Jacob Goldberger.
\newblock Network calibration by class-based temperature scaling.
\newblock In \emph{2021 29th European Signal Processing Conference (EUSIPCO)}, pages 1486--1490. IEEE, 2021.

\bibitem[Geirhos et~al.(2020)Geirhos, Jacobsen, Michaelis, Zemel, Brendel, Bethge, and Wichmann]{geirhos2020shortcut}
Robert Geirhos, J{\"o}rn-Henrik Jacobsen, Claudio Michaelis, Richard Zemel, Wieland Brendel, Matthias Bethge, and Felix~A Wichmann.
\newblock Shortcut learning in deep neural networks.
\newblock \emph{Nature Machine Intelligence}, 2\penalty0 (11):\penalty0 665--673, 2020.

\bibitem[Ghosh et~al.(2024)Ghosh, Evuru, Kumar, Tyagi, Sakshi, Chowdhury, and Manocha]{ghosh2024aspire}
Sreyan Ghosh, Chandra Kiran~Reddy Evuru, Sonal Kumar, Utkarsh Tyagi, S Sakshi, Sanjoy Chowdhury, and Dinesh Manocha.
\newblock Aspire: Language-guided data augmentation for improving robustness against spurious correlations.
\newblock In \emph{Findings of the Association for Computational Linguistics ACL 2024}, pages 386--406, 2024.

\bibitem[Guo et~al.(2017)Guo, Pleiss, Sun, and Weinberger]{guo2017calibration}
Chuan Guo, Geoff Pleiss, Yu Sun, and Kilian~Q Weinberger.
\newblock On calibration of modern neural networks.
\newblock In \emph{International conference on machine learning}, pages 1321--1330. PMLR, 2017.

\bibitem[Huang et~al.(2022)Huang, Wang, Huang, Lee, and Xing]{huang2022two}
Zeyi Huang, Haohan Wang, Dong Huang, Yong~Jae Lee, and Eric~P Xing.
\newblock The two dimensions of worst-case training and their integrated effect for out-of-domain generalization.
\newblock In \emph{Proceedings of the IEEE/CVF Conference on Computer Vision and Pattern Recognition}, pages 9631--9641, 2022.

\bibitem[Ke et~al.(2024)Ke, Ye, Danelljan, Tai, Tang, Yu, et~al.]{ke2024segment}
Lei Ke, Mingqiao Ye, Martin Danelljan, Yu-Wing Tai, Chi-Keung Tang, Fisher Yu, et~al.
\newblock Segment anything in high quality.
\newblock \emph{Advances in Neural Information Processing Systems}, 36, 2024.

\bibitem[Khosla et~al.(2011)Khosla, Jayadevaprakash, Yao, and Fei-Fei]{KhoslaYaoJayadevaprakashFeiFei_FGVC2011}
Aditya Khosla, Nityananda Jayadevaprakash, Bangpeng Yao, and Li Fei-Fei.
\newblock Novel dataset for fine-grained image categorization.
\newblock In \emph{First Workshop on Fine-Grained Visual Categorization, IEEE Conference on Computer Vision and Pattern Recognition}, Colorado Springs, CO, 2011.

\bibitem[Kirillov et~al.(2023)Kirillov, Mintun, Ravi, Mao, Rolland, Gustafson, Xiao, Whitehead, Berg, Lo, et~al.]{kirillov2023segment}
Alexander Kirillov, Eric Mintun, Nikhila Ravi, Hanzi Mao, Chloe Rolland, Laura Gustafson, Tete Xiao, Spencer Whitehead, Alexander~C Berg, Wan-Yen Lo, et~al.
\newblock Segment anything.
\newblock In \emph{Proceedings of the IEEE/CVF International Conference on Computer Vision}, pages 4015--4026, 2023.

\bibitem[Kisel et~al.(2024)Kisel, Volkov, Hanzelkova, Janouskova, and Matas]{kisel2024flaws}
Nikita Kisel, Illia Volkov, Katerina Hanzelkova, Klara Janouskova, and Jiri Matas.
\newblock Flaws of imagenet, computer vision's favourite dataset.
\newblock \emph{arXiv preprint arXiv:2412.00076}, 2024.

\bibitem[Krueger et~al.(2021)Krueger, Caballero, Jacobsen, Zhang, Binas, Zhang, Le~Priol, and Courville]{krueger2021out}
David Krueger, Ethan Caballero, Joern-Henrik Jacobsen, Amy Zhang, Jonathan Binas, Dinghuai Zhang, Remi Le~Priol, and Aaron Courville.
\newblock Out-of-distribution generalization via risk extrapolation (rex).
\newblock In \emph{International conference on machine learning}, pages 5815--5826. PMLR, 2021.

\bibitem[Li et~al.(2018)Li, Pan, Wang, and Kot]{li2018domain}
Haoliang Li, Sinno~Jialin Pan, Shiqi Wang, and Alex~C Kot.
\newblock Domain generalization with adversarial feature learning.
\newblock In \emph{Proceedings of the IEEE conference on computer vision and pattern recognition}, pages 5400--5409, 2018.

\bibitem[Liu et~al.(2021)Liu, Haghgoo, Chen, Raghunathan, Koh, Sagawa, Liang, and Finn]{liu2021just}
Evan~Z Liu, Behzad Haghgoo, Annie~S Chen, Aditi Raghunathan, Pang~Wei Koh, Shiori Sagawa, Percy Liang, and Chelsea Finn.
\newblock Just train twice: Improving group robustness without training group information.
\newblock In \emph{International Conference on Machine Learning}, pages 6781--6792. PMLR, 2021.

\bibitem[Liu et~al.(2023)Liu, Zeng, Ren, Li, Zhang, Yang, Li, Yang, Su, Zhu, et~al.]{liu2023grounding}
Shilong Liu, Zhaoyang Zeng, Tianhe Ren, Feng Li, Hao Zhang, Jie Yang, Chunyuan Li, Jianwei Yang, Hang Su, Jun Zhu, et~al.
\newblock Grounding dino: Marrying dino with grounded pre-training for open-set object detection.
\newblock \emph{arXiv preprint arXiv:2303.05499}, 2023.

\bibitem[Liu et~al.(2022)Liu, Mao, Wu, Feichtenhofer, Darrell, and Xie]{liu2022convnet}
Zhuang Liu, Hanzi Mao, Chao-Yuan Wu, Christoph Feichtenhofer, Trevor Darrell, and Saining Xie.
\newblock A convnet for the 2020s.
\newblock \emph{Proceedings of the IEEE/CVF Conference on Computer Vision and Pattern Recognition (CVPR)}, 2022.

\bibitem[Luo et~al.(2023)Luo, Pan, Sun, Zhang, Xiong, and Wu]{luo2023camouflaged}
Naisong Luo, Yuwen Pan, Rui Sun, Tianzhu Zhang, Zhiwei Xiong, and Feng Wu.
\newblock Camouflaged instance segmentation via explicit de-camouflaging.
\newblock In \emph{Proceedings of the IEEE/CVF conference on computer vision and pattern recognition}, pages 17918--17927, 2023.

\bibitem[Lynch et~al.(2023)Lynch, Dovonon, Kaddour, and Silva]{lynch2023spawrious}
Aengus Lynch, Gbètondji J-S Dovonon, Jean Kaddour, and Ricardo Silva.
\newblock Spawrious: A benchmark for fine control of spurious correlation biases, 2023.

\bibitem[Minderer et~al.(2022)Minderer, Gritsenko, Stone, Neumann, Weissenborn, Dosovitskiy, Mahendran, Arnab, Dehghani, Shen, et~al.]{minderer2022simple}
Matthias Minderer, Alexey Gritsenko, Austin Stone, Maxim Neumann, Dirk Weissenborn, Alexey Dosovitskiy, Aravindh Mahendran, Anurag Arnab, Mostafa Dehghani, Zhuoran Shen, et~al.
\newblock Simple open-vocabulary object detection.
\newblock In \emph{European Conference on Computer Vision}, pages 728--755. Springer, 2022.

\bibitem[Minderer et~al.(2024)Minderer, Gritsenko, and Houlsby]{minderer2024scaling}
Matthias Minderer, Alexey Gritsenko, and Neil Houlsby.
\newblock Scaling open-vocabulary object detection.
\newblock \emph{Advances in Neural Information Processing Systems}, 36, 2024.

\bibitem[Moayeri et~al.(2022{\natexlab{a}})Moayeri, Pope, Balaji, and Feizi]{moayeri2022comprehensive}
Mazda Moayeri, Phillip Pope, Yogesh Balaji, and Soheil Feizi.
\newblock A comprehensive study of image classification model sensitivity to foregrounds, backgrounds, and visual attributes.
\newblock In \emph{Proceedings of the IEEE/CVF Conference on Computer Vision and Pattern Recognition}, pages 19087--19097, 2022{\natexlab{a}}.

\bibitem[Moayeri et~al.(2022{\natexlab{b}})Moayeri, Singla, and Feizi]{moayeri2022hard}
Mazda Moayeri, Sahil Singla, and Soheil Feizi.
\newblock Hard imagenet: Segmentations for objects with strong spurious cues.
\newblock \emph{Advances in Neural Information Processing Systems}, 35:\penalty0 10068--10077, 2022{\natexlab{b}}.

\bibitem[Naeini et~al.(2015)Naeini, Cooper, and Hauskrecht]{naeini2015obtaining}
Mahdi~Pakdaman Naeini, Gregory Cooper, and Milos Hauskrecht.
\newblock Obtaining well calibrated probabilities using bayesian binning.
\newblock In \emph{Proceedings of the AAAI conference on artificial intelligence}, 2015.

\bibitem[Oliva and Torralba(2007)]{oliva2007role}
Aude Oliva and Antonio Torralba.
\newblock The role of context in object recognition.
\newblock \emph{Trends in cognitive sciences}, 11\penalty0 (12):\penalty0 520--527, 2007.

\bibitem[Oquab et~al.(2023)Oquab, Darcet, Moutakanni, Vo, Szafraniec, Khalidov, Fernandez, Haziza, Massa, El-Nouby, et~al.]{oquab2023dinov2}
Maxime Oquab, Timoth{\'e}e Darcet, Th{\'e}o Moutakanni, Huy Vo, Marc Szafraniec, Vasil Khalidov, Pierre Fernandez, Daniel Haziza, Francisco Massa, Alaaeldin El-Nouby, et~al.
\newblock Dinov2: Learning robust visual features without supervision.
\newblock \emph{arXiv preprint arXiv:2304.07193}, 2023.

\bibitem[Picek et~al.(2022)Picek, \v{S}ulc, Matas, Jeppesen, Heilmann-Clausen, L{\ae}ss{\o}e, and Fr{\o}slev]{Picek_2022_WACV}
Luk\'a\v{s} Picek, Milan \v{S}ulc, Ji\v{r}{\'\i} Matas, Thomas~S. Jeppesen, Jacob Heilmann-Clausen, Thomas L{\ae}ss{\o}e, and Tobias Fr{\o}slev.
\newblock Danish fungi 2020 - not just another image recognition dataset.
\newblock In \emph{Proceedings of the IEEE/CVF Winter Conference on Applications of Computer Vision (WACV)}, pages 1525--1535, 2022.

\bibitem[Picek et~al.(2024{\natexlab{a}})Picek, Janouskova, Sulc, and Matas]{picek2024fungitastic}
Lukas Picek, Klara Janouskova, Milan Sulc, and Jiri Matas.
\newblock Fungitastic: A multi-modal dataset and benchmark for image categorization.
\newblock \emph{arXiv preprint arXiv:2408.13632}, 2024{\natexlab{a}}.

\bibitem[Picek et~al.(2024{\natexlab{b}})Picek, Neumann, and Matas]{picek2024animal}
Lukas Picek, Lukas Neumann, and Jiri Matas.
\newblock Animal identification with independent foreground and background modeling.
\newblock \emph{arXiv preprint arXiv:2408.12930}, 2024{\natexlab{b}}.

\bibitem[Radford et~al.(2021)Radford, Kim, Hallacy, Ramesh, Goh, Agarwal, Sastry, Askell, Mishkin, Clark, et~al.]{radford2021learning}
Alec Radford, Jong~Wook Kim, Chris Hallacy, Aditya Ramesh, Gabriel Goh, Sandhini Agarwal, Girish Sastry, Amanda Askell, Pamela Mishkin, Jack Clark, et~al.
\newblock Learning transferable visual models from natural language supervision.
\newblock In \emph{International conference on machine learning}, pages 8748--8763. PMLR, 2021.

\bibitem[Ravi et~al.(2024)Ravi, Gabeur, Hu, Hu, Ryali, Ma, Khedr, R{\"a}dle, Rolland, Gustafson, et~al.]{ravi2024sam}
Nikhila Ravi, Valentin Gabeur, Yuan-Ting Hu, Ronghang Hu, Chaitanya Ryali, Tengyu Ma, Haitham Khedr, Roman R{\"a}dle, Chloe Rolland, Laura Gustafson, et~al.
\newblock Sam 2: Segment anything in images and videos.
\newblock \emph{arXiv preprint arXiv:2408.00714}, 2024.

\bibitem[Rombach et~al.(2022)Rombach, Blattmann, Lorenz, Esser, and Ommer]{rombach2022high}
Robin Rombach, Andreas Blattmann, Dominik Lorenz, Patrick Esser, and Bj{\"o}rn Ommer.
\newblock High-resolution image synthesis with latent diffusion models.
\newblock In \emph{Proceedings of the IEEE/CVF conference on computer vision and pattern recognition}, pages 10684--10695, 2022.

\bibitem[Russakovsky et~al.(2015)Russakovsky, Deng, Su, Krause, Satheesh, Ma, Huang, Karpathy, Khosla, Bernstein, et~al.]{russakovsky2015imagenet}
Olga Russakovsky, Jia Deng, Hao Su, Jonathan Krause, Sanjeev Satheesh, Sean Ma, Zhiheng Huang, Andrej Karpathy, Aditya Khosla, Michael Bernstein, et~al.
\newblock Imagenet large scale visual recognition challenge.
\newblock \emph{International journal of computer vision}, 115:\penalty0 211--252, 2015.

\bibitem[Sagawa et~al.(2019)Sagawa, Koh, Hashimoto, and Liang]{sagawa2019distributionally}
Shiori Sagawa, Pang~Wei Koh, Tatsunori~B Hashimoto, and Percy Liang.
\newblock Distributionally robust neural networks for group shifts: On the importance of regularization for worst-case generalization.
\newblock \emph{arXiv preprint arXiv:1911.08731}, 2019.

\bibitem[Shetty et~al.(2019)Shetty, Schiele, and Fritz]{shetty2019not}
Rakshith Shetty, Bernt Schiele, and Mario Fritz.
\newblock Not using the car to see the sidewalk--quantifying and controlling the effects of context in classification and segmentation.
\newblock In \emph{Proceedings of the IEEE/CVF Conference on Computer Vision and Pattern Recognition}, pages 8218--8226, 2019.

\bibitem[Shi et~al.(2021)Shi, Seely, Torr, Siddharth, Hannun, Usunier, and Synnaeve]{shi2021gradient}
Yuge Shi, Jeffrey Seely, Philip~HS Torr, N Siddharth, Awni Hannun, Nicolas Usunier, and Gabriel Synnaeve.
\newblock Gradient matching for domain generalization.
\newblock \emph{arXiv preprint arXiv:2104.09937}, 2021.

\bibitem[Singla and Feizi(2022)]{singla2022salient}
Sahil Singla and Soheil Feizi.
\newblock Salient imagenet: How to discover spurious features in deep learning?
\newblock In \emph{International Conference on Learning Representations}, 2022.

\bibitem[Stevens et~al.(2024)Stevens, Wu, Thompson, Campolongo, Song, Carlyn, Dong, Dahdul, Stewart, Berger-Wolf, et~al.]{stevens2024bioclip}
Samuel Stevens, Jiaman Wu, Matthew~J Thompson, Elizabeth~G Campolongo, Chan~Hee Song, David~Edward Carlyn, Li Dong, Wasila~M Dahdul, Charles Stewart, Tanya Berger-Wolf, et~al.
\newblock Bioclip: A vision foundation model for the tree of life.
\newblock In \emph{Proceedings of the IEEE/CVF conference on computer vision and pattern recognition}, pages 19412--19424, 2024.

\bibitem[Sun and Saenko(2016)]{sun2016deep}
Baochen Sun and Kate Saenko.
\newblock Deep coral: Correlation alignment for deep domain adaptation.
\newblock In \emph{Computer Vision--ECCV 2016 Workshops: Amsterdam, The Netherlands, October 8-10 and 15-16, 2016, Proceedings, Part III 14}, pages 443--450. Springer, 2016.

\bibitem[Taesiri et~al.(2024)Taesiri, Nguyen, Habchi, Bezemer, and Nguyen]{taesiri2024imagenet}
Mohammad~Reza Taesiri, Giang Nguyen, Sarra Habchi, Cor-Paul Bezemer, and Anh Nguyen.
\newblock Imagenet-hard: The hardest images remaining from a study of the power of zoom and spatial biases in image classification.
\newblock \emph{Advances in Neural Information Processing Systems}, 36, 2024.

\bibitem[Torralba(2003)]{torralba2003contextual}
Antonio Torralba.
\newblock Contextual priming for object detection.
\newblock \emph{International journal of computer vision}, 53:\penalty0 169--191, 2003.

\bibitem[Tschannen et~al.(2025)Tschannen, Gritsenko, Wang, Naeem, Alabdulmohsin, Parthasarathy, Evans, Beyer, Xia, Mustafa, et~al.]{tschannen2025siglip}
Michael Tschannen, Alexey Gritsenko, Xiao Wang, Muhammad~Ferjad Naeem, Ibrahim Alabdulmohsin, Nikhil Parthasarathy, Talfan Evans, Lucas Beyer, Ye Xia, Basil Mustafa, et~al.
\newblock Siglip 2: Multilingual vision-language encoders with improved semantic understanding, localization, and dense features.
\newblock \emph{arXiv preprint arXiv:2502.14786}, 2025.

\bibitem[Vapnik(1991)]{vapnik1991principles}
Vladimir Vapnik.
\newblock Principles of risk minimization for learning theory.
\newblock \emph{Advances in neural information processing systems}, 4, 1991.

\bibitem[Vasudevan et~al.()Vasudevan, Caine, Gontijo-Lopes, Fridovich-Keil, and Roelofs]{VasudevanWhenImageNet}
Vijay Vasudevan, Benjamin Caine, Raphael Gontijo-Lopes, Sara Fridovich-Keil, and Rebecca Roelofs.
\newblock {When does dough become a bagel? Analyzing the remaining mistakes on ImageNet}.

\bibitem[Wang et~al.(2022)Wang, Machiraju, Choung, Herzog, and Frossard]{wang2022clad}
Ke Wang, Harshitha Machiraju, Oh-Hyeon Choung, Michael Herzog, and Pascal Frossard.
\newblock Clad: A contrastive learning based approach for background debiasing.
\newblock \emph{arXiv preprint arXiv:2210.02748}, 2022.

\bibitem[Wang et~al.(2025)Wang, Lin, Chen, Schmidt, Han, and Zhang]{wang2025sober}
Qizhou Wang, Yong Lin, Yongqiang Chen, Ludwig Schmidt, Bo Han, and Tong Zhang.
\newblock A sober look at the robustness of clips to spurious features.
\newblock \emph{Advances in Neural Information Processing Systems}, 37:\penalty0 122484--122523, 2025.

\bibitem[Wightman(2019)]{rw2019timm}
Ross Wightman.
\newblock Pytorch image models.
\newblock \url{https://github.com/rwightman/pytorch-image-models}, 2019.

\bibitem[Woo et~al.(2023)Woo, Debnath, Hu, Chen, Liu, Kweon, and Xie]{woo2023convnext}
Sanghyun Woo, Shoubhik Debnath, Ronghang Hu, Xinlei Chen, Zhuang Liu, In~So Kweon, and Saining Xie.
\newblock Convnext v2: Co-designing and scaling convnets with masked autoencoders.
\newblock In \emph{Proceedings of the IEEE/CVF Conference on Computer Vision and Pattern Recognition}, pages 16133--16142, 2023.

\bibitem[Xiao et~al.(2020)Xiao, Engstrom, Ilyas, and Madry]{xiao2020noise}
Kai Xiao, Logan Engstrom, Andrew Ilyas, and Aleksander Madry.
\newblock Noise or signal: The role of image backgrounds in object recognition.
\newblock \emph{arXiv preprint arXiv:2006.09994}, 2020.

\bibitem[Xu et~al.(2020)Xu, Zhang, Ni, Li, Wang, Tian, and Zhang]{xu2020adversarial}
Minghao Xu, Jian Zhang, Bingbing Ni, Teng Li, Chengjie Wang, Qi Tian, and Wenjun Zhang.
\newblock Adversarial domain adaptation with domain mixup.
\newblock In \emph{Proceedings of the AAAI conference on artificial intelligence}, pages 6502--6509, 2020.

\bibitem[Yang et~al.(2024)Yang, Yang, Wu, and Feng]{yang2024significant}
Shengying Yang, Xinqi Yang, Jianfeng Wu, and Boyang Feng.
\newblock Significant feature suppression and cross-feature fusion networks for fine-grained visual classification.
\newblock \emph{Scientific Reports}, 14\penalty0 (1):\penalty0 24051, 2024.

\bibitem[Yao et~al.(2022)Yao, Wang, Li, Zhang, Liang, Zou, and Finn]{yao2022improving}
Huaxiu Yao, Yu Wang, Sai Li, Linjun Zhang, Weixin Liang, James Zou, and Chelsea Finn.
\newblock Improving out-of-distribution robustness via selective augmentation.
\newblock In \emph{International Conference on Machine Learning}, pages 25407--25437. PMLR, 2022.

\bibitem[Zhao et~al.(2023)Zhao, Ding, An, Du, Yu, Li, Tang, and Wang]{zhao2023fast}
Xu Zhao, Wenchao Ding, Yongqi An, Yinglong Du, Tao Yu, Min Li, Ming Tang, and Jinqiao Wang.
\newblock Fast segment anything.
\newblock \emph{arXiv preprint arXiv:2306.12156}, 2023.

\bibitem[Zhu et~al.(2016)Zhu, Xie, and Yuille]{zhu2016object}
Zhuotun Zhu, Lingxi Xie, and Alan~L Yuille.
\newblock Object recognition with and without objects.
\newblock \emph{arXiv preprint arXiv:1611.06596}, 2016.

\bibitem[Zitnick and Doll{\'a}r(2014)]{zitnick2014edge}
C~Lawrence Zitnick and Piotr Doll{\'a}r.
\newblock Edge boxes: Locating object proposals from edges.
\newblock In \emph{Computer Vision--ECCV 2014: 13th European Conference, Zurich, Switzerland, September 6-12, 2014, Proceedings, Part V 13}, pages 391--405. Springer, 2014.

\end{thebibliography}
}

\clearpage

\appendix

\section{Datasets}
\label{sec:ap:datasets}
\noindent\textbf{ImageNet-1K (IN-1K)} \cite{russakovsky2015imagenet}: A large dataset of diverse 1000 classes, many of which are fine-grained, such as over 100 species of dogs. The dataset features diverse \fg{}-\bg{} relationships. The ImageNet-1K has been tracking the progress of object recognition for more than 10 years. Alongside its wide adoption, it is known to contain many issues~\cite{VasudevanWhenImageNet,Beyer2020AreImageNet,kisel2024flaws}. Recently, a unification of available error corrections was published \cite{kisel2024flaws}. Based on these unified labels, alongside the original dataset, we also evaluate on a \textbf{`clean labels'} subset which only contains images where previous works correcting the dataset agree on the label.

\noindent\textbf{Hard ImageNet (HIN)}
The Hard ImageNet dataset \cite{moayeri2022hard}
is a subset of 15 ImageNet-1K classes \cite{deng2009imagenet}
with strong \fg{}--\bg{} correlations, as observed in
Single et al. \cite{singla2022salient}.
GT segmentation masks are collected using Amazon Mechanical Turk.
The objects in this dataset are less centered and the area they cover is below average.
Of the $ \approx 19000$ training images we reserve $ 10\%$ from each class for validation.
The test set consists of $750$ images. 

For the purpose of assessing model robustness, we introduce two new test sets for this data set.

\noindent\textbf{Hard ImageNet - Long Tail (LT)} contains $226$ images with unusual \fg{}-\bg{} combinations, such as ``volleyball on snow''.

\begin{figure}[tbh]
    \centering
            \setlength{\tabcolsep}{1pt}
    \begin{tabular}{ccc}
        \includegraphics[width=0.32\linewidth]{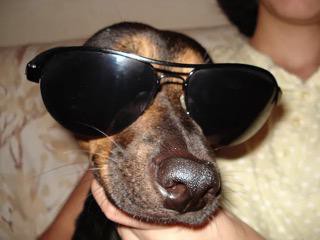} &
        \includegraphics[width=0.32\linewidth]{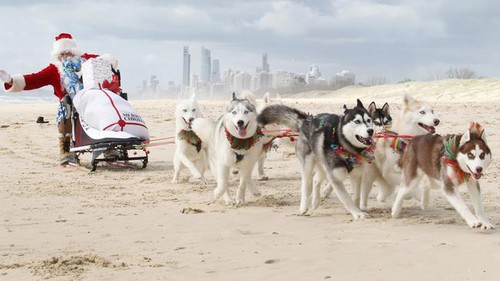} &
        \includegraphics[width=0.32\linewidth]{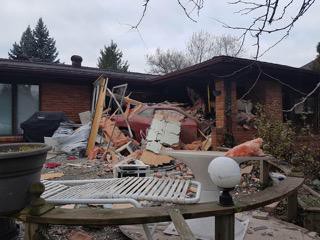} \\
        \includegraphics[width=0.32\linewidth]{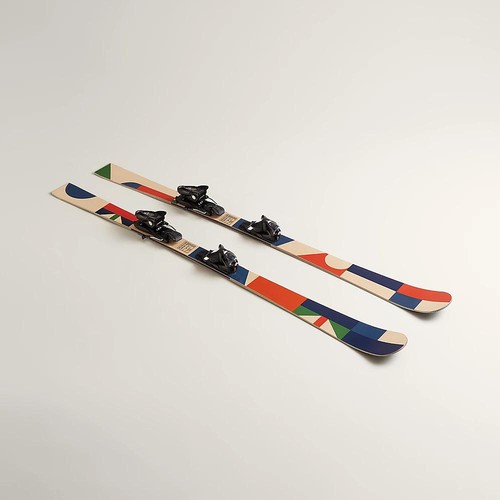} &
        \includegraphics[width=0.32\linewidth]{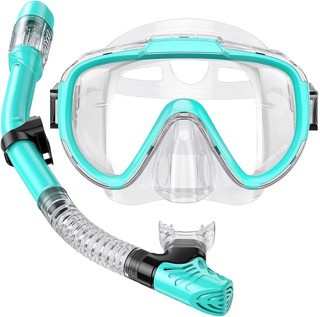} &
        \includegraphics[width=0.32\linewidth]{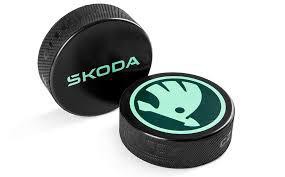} \\
    \end{tabular}
    \caption{Images from the two new test sets for Hard ImageNet - Long Tail (top) and Constant background (bottom).}
    \label{tab:HIN:newimages}
\end{figure}

\noindent\textbf{Hard ImageNet - Constant (CT)} contains $99$ images of essentially constant \bg{}s (commonly co-occurring objects may still appear in the \bg{}, such as snorkel and snorkel mask). See Figure \ref{tab:HIN:newimages} for example images.

\paragraph{FungiTastic}
FungiTastic \cite{picek2024fungitastic} is a challenging fine-grained unbalanced fungi species dataset with complex \fg{}-\bg{} relationships and naturally shifting distribution. In this paper we use the FungiTastic–Mini version of the dataset, where the train set contains observations collected until the end of $ 2021 $ ($46842$ images, $215$ species), while the validation and test sets consist of observations from $ 2022 $ ($9450$ images, $196$ species) and $ 2023 $ ($10914$ images, $193$ species), respectively.
For rare species, only few samples were collected in the training set, and may be missing in either validation or test set. 

The dataset images are accompanied by diverse metadata such as time, GPS location, habitat, substrate, EXIF or toxicity level. The time, substrate and habitat attributes are used to estimate the class priors in some of our experiments.

\noindent\textbf{Spawrious}
\label{subsec:spaw}
The Spawrious datasets \cite{lynch2023spawrious} consist of images with strong \fg{}-\bg{} correlations generated using \emph{Stable Diffusion v1.4} \cite{rombach2022high}. We demonstrate our method on the O2O-E Env1 dataset, where each of the $ 4 $ dog breed classes is associated with one of the $ 4 $ different backgrounds (\emph{Desert}, \emph{Jungle}, \emph{Dirt}, \emph{Snow}) in the training set.

The \bg{}s are permuted in the test set, creating a significant domain shift. 

Two variations of the Spawrious dataset are analyzed: 
\begin{itemize}
    \item For the Resnet-50 experiments in Tables \ref{table:spaw:resnetfull} and \ref{tab:spaw:r50}  
 we follow the process of \cite{lynch2023spawrious} in which training set \fg{}-\bg{} combinations are set to $ 97 \%$
(e.g. \emph{Bulldogs} appear in \emph{Desert} $ 97 \%$ of the time, and on \emph{Beach} $ 3 \%$ of the time ), while test set images for one class always contain the same \bg{} (test \emph{Bulldogs} always appear on \emph{Dirt} \bg{}). See \cite{lynch2023spawrious}[Table 2] for more details. 
\item For the rest of our Spawrious experiments, including Tables \ref{tab:summary} and 
\ref{tab:spaw:fullres}, the training correlations are set to $100 \%$ as well. 
\end{itemize}

Of the $ 12672 $ images in the initial training set, $ 10 \% $ are reserved for validation.

\noindent\textbf{Stanford Dogs}
The Stanford Dogs dataset \cite{KhoslaYaoJayadevaprakashFeiFei_FGVC2011}, a curated subset of \cite{deng2009imagenet}, contains $ 20,580 $ images of dogs from around the world belonging to $ 120 $ species, with $ \approx 150-200 $ images per class. A large portion of settings are in man-made environments, resulting in larger \bg{} variation compared to other animal datasets. There are no strong \fg{}-\bg{} correlations that we are aware of.

\noindent\textbf{Counter Animal} A dataset of 45 animal classes from IN-1K with images from the iNaturalist collection of wildlife images introduced in \cite{wang2025sober}. Each image is further labelled based on the \bg{} as `common' or `rare'. To construct the dataset, the researches first checked CLIP's accuracy on the images and then identified which types of backgrounds are present in images with low accuracy. The construction process disregards the fact that there can be many reasons for the accuracy drop and that \bg{} changes are often highly correlated with a change in appearance - for instance, photos of birds in the sky are typically captured mid-flight, from greater distance than on the ground, and in a very different pose. Often, the animals in the `rare' group are captured in an environment where they are hard to localize even for humans due to camouflage, while the common subset captures them on a white background. Overall, for many classes, it is not clear whether it is the change in \bg{} distribution or the change in \fg{} distribution that causes the accuracy drop.

\begin{figure*}
    \centering
    \begin{tabular}{c}
    \includegraphics[width=0.95\linewidth]{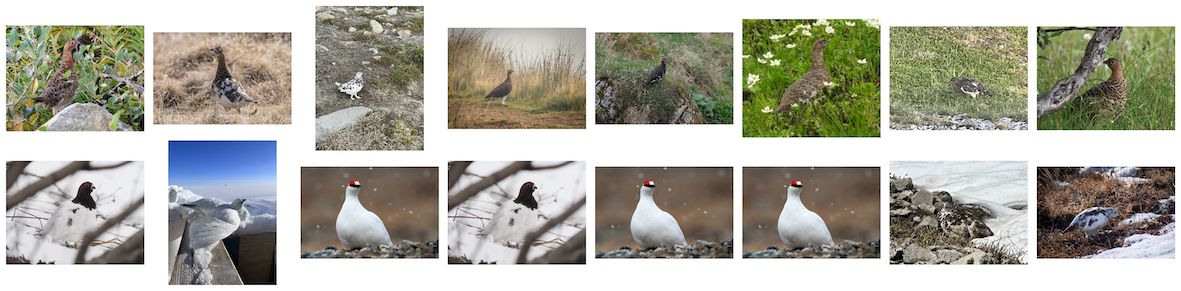} \\
     \includegraphics[width=0.95\linewidth]
     {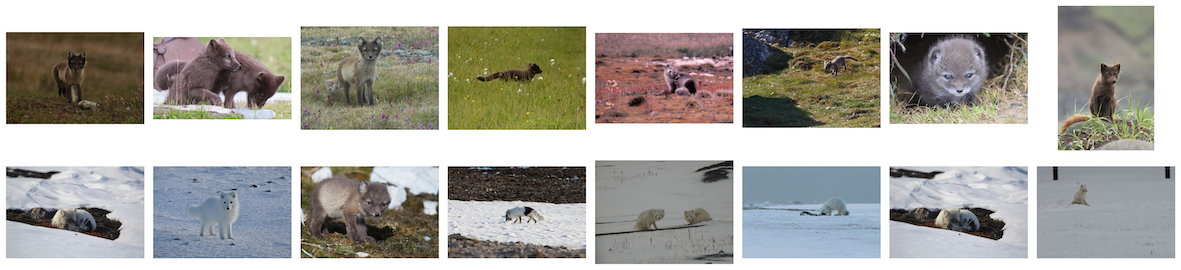} \\
     Animals like polar foxes change appearance between winter (`snow') and summer (`grass'). \\
     
     \includegraphics[width=0.95\linewidth]{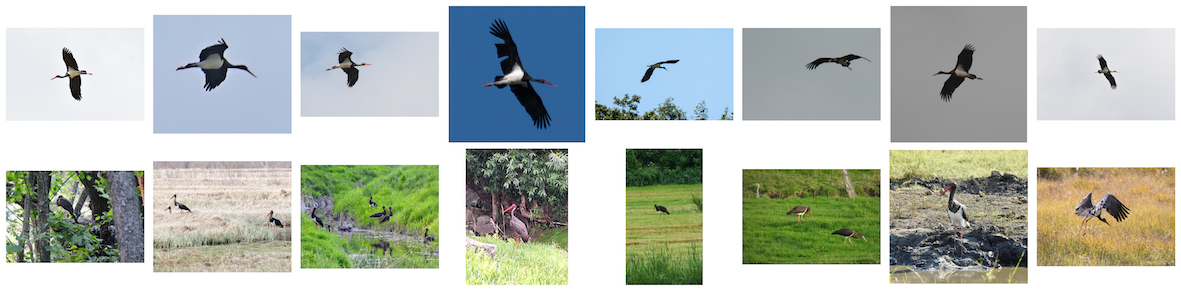} \\
     \includegraphics[width=0.95\linewidth]{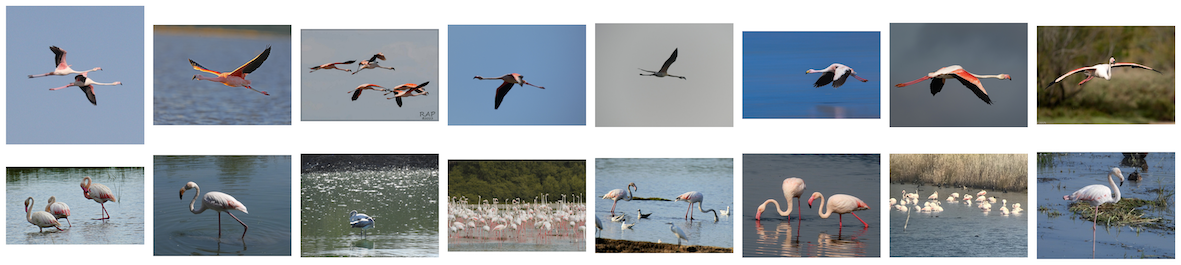} \\
     The pose of flying birds is very different from those on the ground. \\
     
     \includegraphics[width=0.95\linewidth]{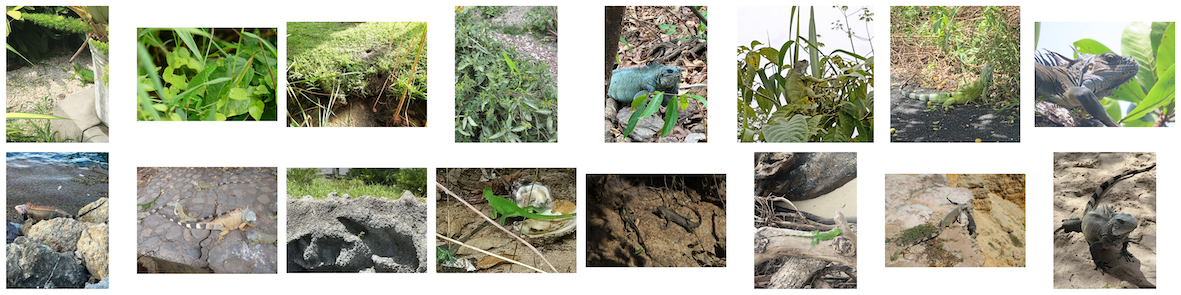} \\
     Green iguanas on a green background, often hidden among leaves, are hard to spot even for a human.

    \end{tabular}
    \caption{Common problems in the counter animal dataset. Each row shows a random sample of images from a class/\bg{} combination. The top row of images always shows images from the 'rare'\bg{} group while the bottom row shows images from the same class, 'common' \bg{} subset.}
    \label{fig:counter_animal_problems}
\end{figure*}

We illustrate some of the issues on randomly sampled images in Figure \ref{fig:counter_animal_problems}. The dataset also contains many duplicate images or images where the animal is not even visible.

\section{Methods}
\label{sec:ap:methods}

\begin{figure*}[bt]
\centering
\includegraphics[width=0.98\linewidth]{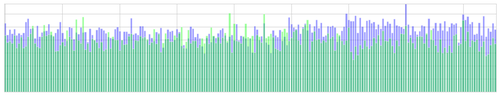}
\caption{
The relative role of {\setlength{\fboxsep}{1pt}\colorbox{blue!30}{\fg{}}} and {\setlength{\fboxsep}{1pt}\colorbox{green!30}{\bg{}}} for the 215 FungiTastic classes shown
by the weights of the learned weighted logits combination model, i.e. Model \ref{ap:weightedLogits}. in  Section \ref{ap:methods:comb}.
The \bg{}s  has a higher weight for about  15\%
}
\label{fig:FungiWeights}
\end{figure*}

\subsection{Localization}

\paragraph{Fine-grained datasets and Counter Animal}
The detctions are produced by the open-set object detector Grounding DINO \cite{liu2023grounding} from dataset-specific text prompts, as discussed in the main text.
For Stanford Dogs and Spawrious datasets, segmentation masks are generated with the text prompt `full dog' while the prompt `mushroom' is used for the FungiTastic. 

For Counter Animal, the prompt is composed as an average of the embeddings of the following meta-prompts: `animal', `bird', `insect', `reptile'. 

\paragraph{Hard ImageNet}

\begin{figure}[bt]
\centering
\includegraphics[width=\linewidth]{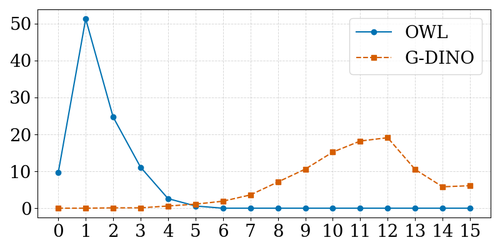}
\caption{Object detection by OWL+SAM and GroundingDino+SAM  on the Hard ImageNet validation set ($\approx 1900$ images). 
For each image, the zero-shot detector is prompted for each class, producing $s \in (0, \cdots 15)$ non-empty 
segmentation masks.
The histogram of the $s$ values is shown.
For example, $51.3 \% $ of images get a single OWL+SAM mask.
The GroundingDINO results show a higher number of masks. The value of $k$ is optimized on the validation set.
} 
\label{fig:detector_fps}
\end{figure}

GroundingDINO works well in cases when it is known a priori that an object matching the text exists in the image. Otherwise, it is prone to output false positives. This is the case for Hard ImageNet, where we prompt an image with texts corresponding to multiple labels. Then, false positive \fg{} outputs (e.g. a person) correlated to a different class (e.g. sunglasses) may confuse the model. To mitigate this, we replace Grounding DINO with the OWLv2 detector \cite{minderer2024scaling, minderer2022simple}, at the expense of introducing more false negatives.  
A comparison of OWLv2 and GroundingDINO in terms of average number of object proposals per image is provided in Figure \ref{fig:detector_fps}.

When a segmentation mask is required, such as for experiments that consider \bg{} with shape inputs, we prompt the SAM \cite{kirillov2023segment} model with the detected bounding boxes.

\subsection{Input options}
\label{sec:input-options}

We extend the different methods to create the $ x, x_{\text{FG}}, x_{\text{BG}} $ images introduced in the main text.
These are input options for the \full{}, \fg{} and \bg{} base classifiers, and represent the rows in Tables \ref{table:fungitastic:full}-\ref{tab:stanfdogs:fullres}. The different options are:
\begin{enumerate}
    \item \full{} images - the standard approach.
    \item \fg{}$_\text{C}$: the image is cropped according to the bounding box and padded to a square to preserve the aspect ratio. 
    \item \fg{}$_\text{M}$: the \bg{} is fully masked before cropping a square bounding box. 
    \item \bg{}$_\text{S}$: \bg{} images with shape (the \fg{} are masked, but their shapes remain)
    \item \bg{}$_\text{B}$: \bg{} w/o shape (a minimal segmentation bounding box masks the \fg{})
\end{enumerate}

\begin{figure*}
    \centering

    \begin{tabular}{ccccc}
        \includegraphics[width=0.18\textwidth, height=2.3cm]{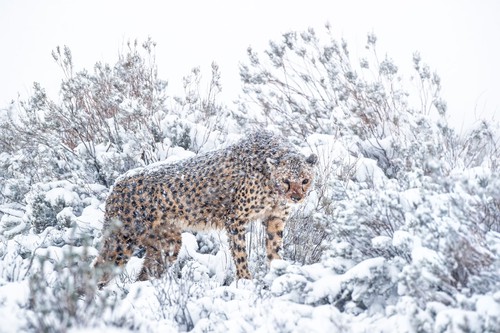} 
        &

        \includegraphics[width=0.16\textwidth, height=2.3cm]{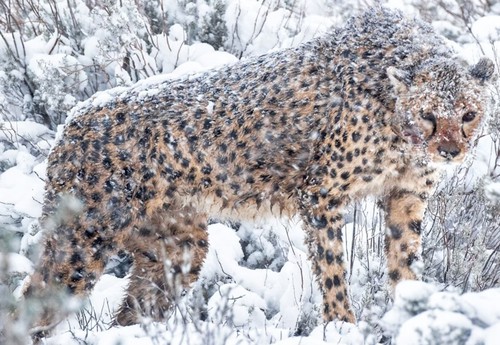}
        
        &
        \includegraphics[width=0.16\textwidth, height=2.3cm]{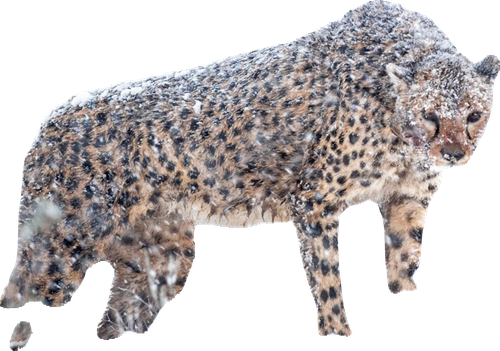}
        
        &
        \includegraphics[width=0.18\textwidth, height=2.3cm]{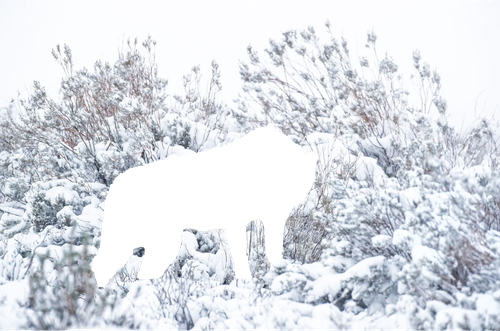}
        &
        \includegraphics[width=0.18\textwidth, height=2.3cm]{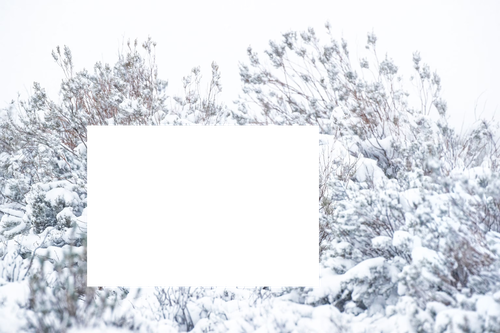}
        \\
        \full{} & \fg{}$_\text{C}$ & \fg{}$_\text{M}$ & \bg{}$_\text{S}$ & \bg{}$_\text{B} $
    \end{tabular}

    \caption{Different kinds of input to the \fg{} and \bg{} models. \full{} is the standard full image, \fg{}$_\text{C}$ is cropped based on the segmentation bounding box, \fg{}$_\text{M}$ is same as \fg{}$_\text{C}$ but with the \bg{} regions masked out, \bg{}$_\text{S}$ is the shape-preserving \bg{} model with \fg{} regions masked out and \bg{}$_\text{B}$ has the area corresponding to the the segmentation bounding box masked out.}
\label{fig:fg_bg_inputs}
\end{figure*}

A visualization is presented in Figure
\ref{fig:fg_bg_inputs}.

\subsection{Combined models} \label{ap:methods:comb}
Here we present the fusion models 
in detail, including the temperature-scaled variants. 

We consider two fixed trained models: $ \Phi_1 $ and $ \Phi_2 $, which output logit vectors $ \Phi_i(x) = z_i \in \mathbb{R}^C $, to which softmax activations are applied: $ 
\sigma(z_i) $. Predictions are obtained by $ \hat{y}_i = \argmax_k z_i^{(k)} $ and their confidences by $ \hat{p}_i  =  \max_k \sigma(z_i)^{(k)} $, \ $ i = 1,2 $. 

\

Since the predictions may be over/under-confident (i.e. the confidences do not reflect the accuracies) and we want to compare the confidences of different models, we opt for calibrating them using the method of \emph{temperature scaling} \cite{guo2017calibration}. This is done using a single parameter $ T >  0 $ for all classes. Given a model  $ \Phi $, the logits and confidences are scaled by 
\begin{equation} \label{tempscale:ap:eq}
z  \to z/T, \ \sigma(z) \to \sigma(z/T), \  \hat{p} \to \tilde{p} =  \max_k \sigma(z/T)^{(k)}
\end{equation}
Note that the predictions of a fixed model do not change, since the same parameter is applied to all classes. This parameter $ T $ is optimized such that the cross entropy loss is minimized on the \emph{validation} set. 

For some datasets it may be desirable to apply different scaling parameters for each class. Such a \emph{class-based temperature scaling} calibration methods was proposed in \cite{frenkel2021network}, attempting to minimize the \emph{expected calibration error} (ECE) \cite{naeini2015obtaining} on the validation set, while not decreasing accuracy, by performing a greedy grid-search. This results in modified logits:
\begin{equation} \label{multitempscale:ap:eq}
z = (z^{(1)}, \dots, z^{(C)}) \to (z^{(1)}/T_1, \dots, z^{(C)}/T_C) \quad 
\end{equation}

\paragraph{Confidence fusion}
\begin{enumerate}
\item (Max confidence) Between $ \hat{y}_1 $ and $ \hat{y}_2 $ choose the most confident prediction $ \hat{y}_i $, i.e. the one with confidence $  \hat{p}_i = \max (\hat{p}_1, \hat{p}_2 ) $.
\item (Max scaled confidence) Again we choose the more confident prediction $ \hat{y}_i $, but now the confidences are calibrated using temperature scaling from Equation \eqref{tempscale:ap:eq}, originating from $ z_1/{T_1} $,   $ z_2/{T_2} $, i.e. choose the one with $  \tilde{p}_i = \max (\tilde{p}_1, \tilde{p}_2 ) $.

\item (Threshold prediction) We choose $ \hat{y}_1 $ if $ \hat{p}_1 > t $, otherwise choose the higher confidence prediction. Here $ t>0 $ is a parameter maximizing the new prediction accuracy on the validation set. 

\item \label{ap:tempscaledavg} (Temperature-scaled average) Let $ z_1/{T_1} $,   $ z_2/{T_2} $ be the scaled logits vectors from Equation \eqref{tempscale:ap:eq} from the two models and take the average $ 
\frac{1}{2} ( \sigma(z_1/{T_1}) + \sigma(z_2/{T_2})) $. The prediction is given by $\argmax $ as usual. 

\item \label{ap:tempscaledWavg} (Temperature-scaled weighted average) As before, but take a weighted average $  \alpha \sigma(z_1/{T_1}) + (1-\alpha) \sigma(z_2/{T_2}) $, where  $ \alpha \in [0,1] $  maximizes  validation set prediction accuracy. 
\end{enumerate}
\paragraph{Learnt fusion}
Finally, the predictions learned from the combined logits on the train set are:
\begin{enumerate}
\setcounter{enumi}{7}
\item \label{ap:concatFC} (Concatenate + FC layers)  To model the interaction between outputs of $ \Phi_1 $ and $ \Phi_2 $, we create new (train, validation and test) datasets by concatenating the logits for each sample $ x $:
\begin{equation} \label{concatlogits:ap:eq}
\begin{split}
{\bf \Psi}(x) = ( \Phi_1(x),  \Phi_2(x)) = (z_1, z_2 ) = \\ 
(z^{(1)}_1, \dots, z^{(C)}_1, z^{(1)}_2, \dots, z^{(C)}_2) \in \mathbb{R}^{2C}
\end{split}
\end{equation}

We input Equation \eqref{concatlogits:ap:eq} into a shallow fully connected network, whose weights are learned from the training set, with cross entropy loss. This can learn more flexible combinations, but it lacks in interpretability and may overfit if the number of classes is large.  

\item \label{ap:weightedLogits} (Weighted logits combination) Generalizes the averages from confidence fusion  by allowing the weights to be class-dependent vectors $ w_1, w_2 \in \mathbb{R}^C $, representing combined logits as $ w_1 z_1 + w_2 z_2 = $
$$
 (w^{(1)}_1 z^{(1)}_1 + w^{(1)}_2 z^{(1)}_2, \dots ,w^{(C)}_1 z^{(C)}_1 + w^{(C)}_2 z^{(C)}_2). 
$$
We optimize the cross entropy loss instead of the ECE from Equation \eqref{multitempscale:ap:eq}, so gradient descent becomes applicable replacing the grid search. We optimize the weights on the training set instead of the validation set. Compared to the FC model \ref{ap:concatFC}, there are much fewer parameters, so there is less risk of overfitting 
and the network weights are more interpretable (see Fig. \ref{fig:FungiWeights}). 

\end{enumerate}

\section{Additional experiments}
\label{sec:ap:exps}

\begin{figure}[bt]
    \centering
        \setlength{\tabcolsep}{3pt}
    \begin{tabular}{c|cc}
         \includegraphics[trim={0cm 5cm 0cm 5cm},clip,width=0.26\linewidth,]{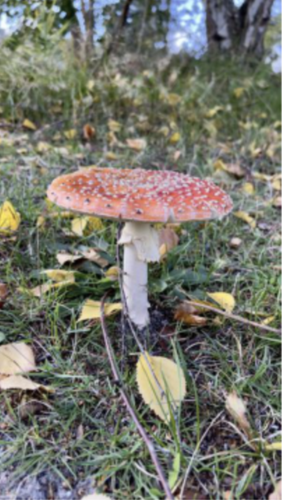}
         & 
    \includegraphics[width=0.31\linewidth,]{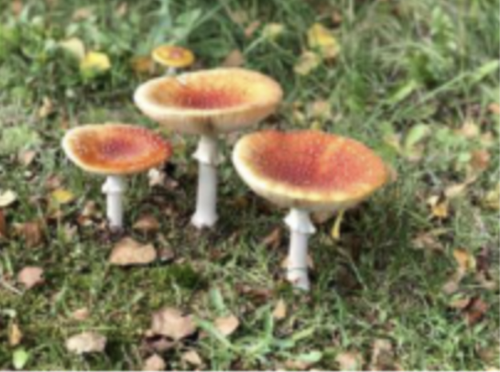} 
         &
         \includegraphics[width=0.31\linewidth,]{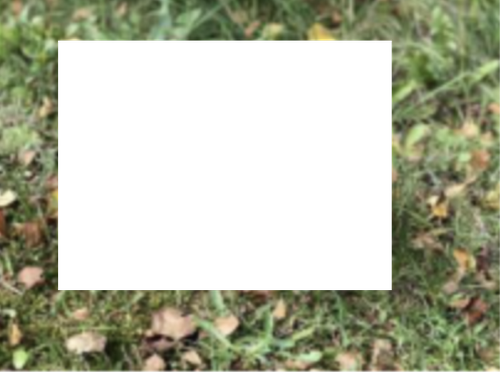}  
         \\

         \includegraphics[trim={0cm 1cm 0cm 1cm},clip,width=0.265\linewidth,] {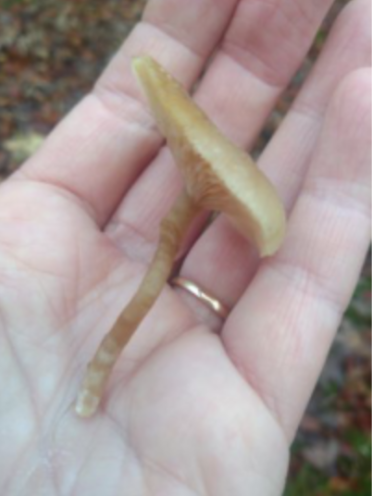}
         & 
        \includegraphics[trim={0cm 1cm 0cm 2cm},clip,width=0.31\linewidth,]{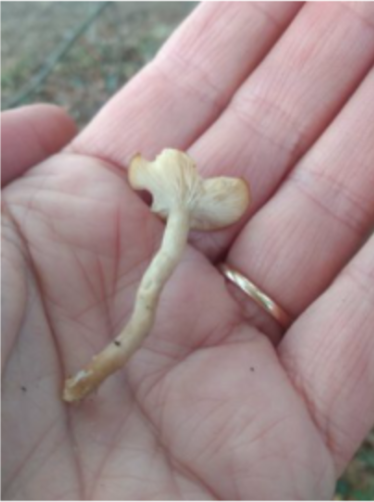} 
         &
         \includegraphics[trim={0cm 1cm 0cm 2cm},clip,width=0.31\linewidth,]{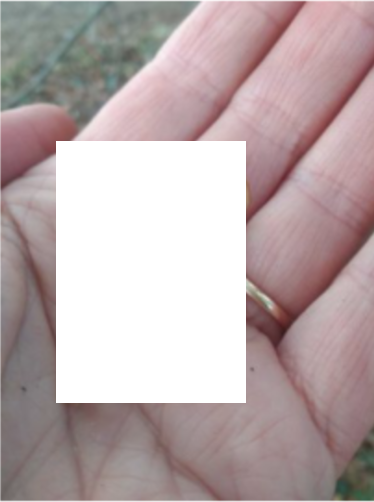}  
         \\
    \end{tabular}
    \caption{
    Extreme \bg{} overfitting on the Fungitastic dataset \cite{picek2024fungitastic}.
     Both the \full{} (middle column, expected) and \bg{} (right column, surprising) images are correctly classified, with high confidence.
     The likely reason is the presence of a unique ``background'' feature, the hand (bottom left) and the existence of a training image (taken in a different year) acquired from a very similar viewpoint (top left).
     The problem highlights the benefits of the interpretability of \strr{}: with uncertain \fg{} prediction, one should not pick and eat the mushroom no matter how confident the \full{} or \bg{} predictions are.
     Note: the mushrooms in each row are the same species.
     }

    \label{fig:fungi_bg_imp}
\end{figure}

\begin{figure}[bt]
\centering
\includegraphics[width=0.98\linewidth]{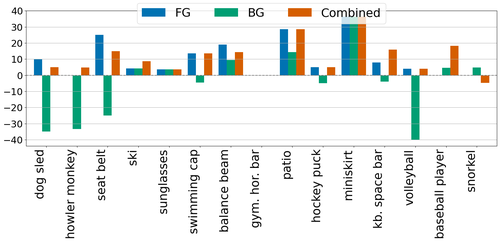}
\caption{
Per-class accuracy \% increase or decrease w.r.t. full-image performance on the proposed HardImageNet test set with a significant \bg{} distribution shift. The experiment confirms that \fg{}$_C$ (cropping image based on segmenation bounding box) is a strong baseline, adding 5.5\% in accuracy. The combined approach performs the best, adding 8 \% of accuracy. Surprisingly,  the \bg{} classifier performs well on about half of the classes, probably due to the information preserved by the mask shape, despite the domain shift.
}
\label{fig:hin_per_class}
\end{figure}

\begin{table*}
\small
\centering
\setlength{\tabcolsep}{3pt}
\begin{tabular}{l rrr r r rr r}
\toprule
 & \multicolumn{3}{c}{\textbf{HardImageNet*}} & \textbf{Dogs} & \textbf{Spaw} & \multicolumn{2}{c}{\textbf{CounterAnimal}} & \textbf{Fungi}  \\
 & Original & CT & LT & Test & Test & Common & Rare & Original \\
\cmidrule(r){2-4} \cmidrule(r){5-5} \cmidrule(r){6-6} \cmidrule(r){7-8}

\multicolumn{9}{l}{\textbf{CLIP-B}} \\
\fg{}$\oplus_{\text{max}}$\bg & {\scriptsize\textcolor{ForestGreen}{+2.4}} 88.93 & {\scriptsize\textcolor{red}{-1.7}} 87.88 & {\scriptsize\textcolor{ForestGreen}{+6.82}} 84.51 & {\scriptsize\textcolor{ForestGreen}{+7.56}} 58.09 & {\scriptsize\textcolor{ForestGreen}{+4.05}} 89.86 & {\scriptsize\textcolor{ForestGreen}{+0.57}} 84.08 & {\scriptsize\textcolor{red}{-0.29}} 67.75 & {\scriptsize\textcolor{ForestGreen}{+1.07}} 2.82 \\
\fg{}$\oplus_{\text{max}}$\full{} & {\scriptsize\textcolor{ForestGreen}{+2.8}} 89.33 & {\scriptsize\textcolor{ForestGreen}{+2.34}} 91.92 & {\scriptsize\textcolor{ForestGreen}{+6.82}} 84.51 & {\scriptsize\textcolor{ForestGreen}{+7.0}} 57.53 & {\scriptsize\textcolor{ForestGreen}{+4.74}} 90.55 & {\scriptsize\textcolor{ForestGreen}{+2.18}} 85.69 & {\scriptsize\textcolor{ForestGreen}{+0.54}} 68.58 & {\scriptsize\textcolor{ForestGreen}{+1.91}} 3.66 \\
Bg & {\scriptsize\textcolor{red}{-5.2}} 81.33 & {\scriptsize\textcolor{red}{-33.71}} 55.87 & {\scriptsize\textcolor{red}{-21.38}} 56.31 & {\scriptsize\textcolor{red}{-45.43}} 5.10 & {\scriptsize\textcolor{red}{-46.83}} 38.98 & {\scriptsize\textcolor{red}{-13.35}} 70.16 & {\scriptsize\textcolor{red}{-15.08}} 52.96 & {\scriptsize\textcolor{red}{-0.74}} 1.01 \\
\fg{} & {\scriptsize\textcolor{red}{-0.53}} 86.00 & {\scriptsize\textcolor{red}{-2.46}} 87.12 & {\scriptsize\textcolor{ForestGreen}{+1.69}} 79.38 & {\scriptsize\textcolor{ForestGreen}{+7.5}} 58.03 & {\scriptsize\textcolor{ForestGreen}{+5.26}} 91.07 & {\scriptsize\textcolor{red}{-1.92}} 81.59 & {\scriptsize\textcolor{ForestGreen}{+0.32}} 68.36 & {\scriptsize\textcolor{ForestGreen}{+0.05}} 1.80 \\
\full{} & 86.53 & 89.58 & 77.69 & 50.53 & 85.81 & 83.51 & 68.04 & 1.75 \\

\midrule

\multicolumn{9}{l}{\textbf{BioCLIP}} \\
\fg{}$\oplus_{\text{max}}$\bg & {\scriptsize \textcolor{red}{-0.4}} 19.60 & {\scriptsize \textcolor{red}{-2.61}} 14.14 & {\scriptsize \textcolor{red}{-1.39}} 15.93 & {\scriptsize \textcolor{red}{-0.39}} 2.84 & {\scriptsize \textcolor{ForestGreen}{+0.10}} 40.83 & {\scriptsize \textcolor{red}{-0.16}} 82.29 & {\scriptsize \textcolor{red}{-2.73}} 75.63 & {\scriptsize \textcolor{ForestGreen}{+14.92}} 33.54 \\
\fg{}$\oplus_{\text{max}}$\full{} & {\scriptsize \textcolor{ForestGreen}{+0.27}} 20.27 & {\scriptsize \textcolor{ForestGreen}{+2.44}} 19.19 & {\scriptsize \textcolor{red}{-2.72}} 14.60 & {\scriptsize \textcolor{red}{-0.18}} 3.05 & {\scriptsize \textcolor{ForestGreen}{+0.91}} 41.64 & {\scriptsize \textcolor{ForestGreen}{+2.27}} 84.72 & {\scriptsize \textcolor{red}{-0.08}} 78.28 & {\scriptsize \textcolor{ForestGreen}{+19.18}} 37.80 \\
\bg{} & {\scriptsize \textcolor{red}{-1.73}} 18.27 & {\scriptsize \textcolor{red}{-9.10}} 7.65 & {\scriptsize \textcolor{red}{-6.99}} 10.33 & {\scriptsize \textcolor{red}{-2.30}} 0.93 & {\scriptsize \textcolor{red}{-3.03}} 37.70 & {\scriptsize \textcolor{red}{-21.25}} 61.20 & {\scriptsize \textcolor{red}{-20.21}} 58.15 & {\scriptsize \textcolor{red}{-16.43}} 2.19 \\
\fg{} & {\scriptsize \textcolor{red}{-5.87}} 14.13 & {\scriptsize \textcolor{ForestGreen}{+0.89}} 17.64 & {\scriptsize \textcolor{red}{-3.69}} 13.63 & {\scriptsize \textcolor{ForestGreen}{+0.03}} 3.26 & {\scriptsize \textcolor{ForestGreen}{+0.92}} 41.65 & {\scriptsize \textcolor{red}{-3.90}} 78.55 & {\scriptsize \textcolor{red}{-4.08}} 74.28 & {\scriptsize \textcolor{red}{-1.75}} 16.87 \\
\full{} & 20.00 & 16.75 & 17.32 & 3.23 & 40.73 & 82.45 & 78.36 & 18.62 \\

\midrule

\multicolumn{9}{l}{\textbf{CLIP-L}} \\
\fg{}$\oplus_{\text{max}}$\bg & {\scriptsize \textcolor{ForestGreen}{+1.34}} 94.27 & {\scriptsize \textcolor{ForestGreen}{+1.41}} 94.95 & {\scriptsize \textcolor{ForestGreen}{+3.52}} 93.36 & {\scriptsize \textcolor{ForestGreen}{+5.10}} 73.25 & {\scriptsize \textcolor{ForestGreen}{+0.15}} 94.89 & {\scriptsize \textcolor{red}{-0.33}} 92.64 & {\scriptsize \textcolor{ForestGreen}{+0.85}} 85.17 & {\scriptsize \textcolor{ForestGreen}{+1.42}} 3.10 \\
\fg{}$\oplus_{\text{max}}$\full{} & {\scriptsize \textcolor{ForestGreen}{+2.40}} 95.33 & {\scriptsize \textcolor{ForestGreen}{+1.41}} 94.95 & {\scriptsize \textcolor{ForestGreen}{+3.97}} 93.81 & {\scriptsize \textcolor{ForestGreen}{+4.62}} 72.77 & {\scriptsize \textcolor{ForestGreen}{+0.69}} 95.43 & {\scriptsize \textcolor{ForestGreen}{+0.38}} 93.35 & {\scriptsize \textcolor{ForestGreen}{+1.12}} 85.44 & {\scriptsize \textcolor{ForestGreen}{+2.26}} 3.94 \\
\bg{} & {\scriptsize \textcolor{red}{-2.53}} 90.40 & {\scriptsize \textcolor{red}{-29.51}} 64.03 & {\scriptsize \textcolor{red}{-16.52}} 73.32 & {\scriptsize \textcolor{red}{-59.15}} 9.00 & {\scriptsize \textcolor{red}{-24.14}} 70.60 & {\scriptsize \textcolor{red}{-15.74}} 77.23 & {\scriptsize \textcolor{red}{-16.55}} 67.77 & {\scriptsize \textcolor{red}{-1.20}} 0.48 \\
\fg{} & {\scriptsize \textcolor{red}{-0.66}} 92.27 & {\scriptsize \textcolor{ForestGreen}{+1.02}} 94.56 & {\scriptsize \textcolor{red}{-1.76}} 88.08 & {\scriptsize \textcolor{ForestGreen}{+5.08}} 73.23 & {\scriptsize \textcolor{ForestGreen}{+0.71}} 95.45 & {\scriptsize \textcolor{red}{-0.99}} 91.98 & {\scriptsize \textcolor{ForestGreen}{+0.23}} 84.55 & {\scriptsize \textcolor{ForestGreen}{+0.11}} 1.79 \\
\full{} & 92.93 & 93.54 & 89.84 & 68.15 & 94.74 & 92.97 & 84.32 & 1.68 \\

\midrule
\multicolumn{9}{l}{\textbf{SigLIP2-SO}} \\
\fg{}$\oplus_{\text{max}}$\bg & {\scriptsize\textcolor{ForestGreen}{+1.34}} 96.67 & {\scriptsize\textcolor{ForestGreen}{+0.00}} 100.00 & {\scriptsize\textcolor{ForestGreen}{+1.57}} 97.79 & {\scriptsize\textcolor{ForestGreen}{+0.58}} 84.69 & {\scriptsize\textcolor{red}{-0.41}} 94.93 & {\scriptsize\textcolor{red}{-0.63}} 94.87 & {\scriptsize\textcolor{red}{-2.61}} 86.75 & --- \\
\fg{}$\oplus_{\text{max}}$\full{} & {\scriptsize\textcolor{ForestGreen}{+1.07}} 96.40 & {\scriptsize\textcolor{ForestGreen}{+0.00}} 100.00 & {\scriptsize\textcolor{ForestGreen}{+2.01}} 98.23 & {\scriptsize\textcolor{ForestGreen}{+0.97}} 85.08 & {\scriptsize\textcolor{ForestGreen}{+1.09}} 96.43 & {\scriptsize\textcolor{ForestGreen}{+0.17}} 95.67 & {\scriptsize\textcolor{red}{-1.16}} 88.20 & --- \\
\bg{} & {\scriptsize\textcolor{red}{-1.86}} 93.47 & {\scriptsize\textcolor{red}{-10.54}} 89.46 & {\scriptsize\textcolor{red}{-11.12}} 85.10 & {\scriptsize\textcolor{red}{-62.37}} 21.74 & {\scriptsize\textcolor{red}{-16.92}} 78.42 & {\scriptsize\textcolor{red}{-11.49}} 84.01 & {\scriptsize\textcolor{red}{-13.47}} 75.89 & --- \\
\fg{} & {\scriptsize\textcolor{red}{-2.53}} 92.80 & {\scriptsize\textcolor{ForestGreen}{+0.00}} 100.00 & {\scriptsize\textcolor{red}{-1.67}} 94.55 & {\scriptsize\textcolor{ForestGreen}{+0.37}} 84.48 & {\scriptsize\textcolor{ForestGreen}{+1.33}} 96.67 & {\scriptsize\textcolor{red}{-1.31}} 94.19 & {\scriptsize\textcolor{red}{-0.92}} 88.44 & --- \\
\full{} & 95.33 & 100.00 & 96.22 & 84.11 & 95.34 & 95.50 & 89.36 & --- \\
\bottomrule
\end{tabular}
\caption{Performance of VLM models on different kinds of inputs - \full{}, \fg{} and \bg{}. Maximum confidence fusion of \fg{} + \bg{}, as well as \fg{} + \full{}, is also reported. --- SigLIP2 results were left out due to the high GPU memory requirements on datasets with a high number of classes. *Resutls obtained with oracle detections generated by GT prompting.
}
\label{tab:vlm_comp_full}
\end{table*}

\begin{table}[tbh]
    \centering
    \setlength{\tabcolsep}{5pt}

    \begin{tabular}{llccc}
        \toprule
        & & \multicolumn{3}{c}{\textbf{HardImageNet Test Sets}} \\
        \cmidrule{3-5}
        & \textbf{Model} & \textbf{Original} & \textbf{Long Tail} & \textbf{Constant} \\
        \midrule
        & \full{} & 97.33 & 81.33 & 90.51 \\
        \midrule
        
        \multirow{3}{*}{\rotatebox{90}{\small GT masks}} 
        & \fg{} & 97.63 & 87.08 & 94.55 \\
        & \bg{} & 95.01 & 78.94 & 67.88 \\
        & \fg{}$\oplus$\bg{} & 98.45 & 89.56 & 90.30 \\
        \midrule
        
        \multirow{3}{*}{\rotatebox{90}{\small GT labels}} 
        & \fg{} & 97.79 & 85.93 & 90.10 \\
        & \bg{} & 97.84 & 79.73 & 73.94 \\
        & \fg{}$\oplus$\bg{} & 98.99 & 90.0 & 92.12 \\
        \midrule
        
        \multirow{4}{*}{\rotatebox{90}{\small No GT}} 
        & \fg{} &95.55 & 81.24 & 90.1 \\
        & \bg{} & 96.83 & 80.27 & 88.28 \\
        & \full{}$\oplus$\fg{} & 97.68 & 90.91 & 83.45  \\
        & \full{}$\oplus$\fg{}$\oplus$\bg{} & 98.03 & 83.63 & 91.31 
        \\ 
        & \full{}$\oplus_{*}$\full{} & 97.51 & 90.91 & 82.89 \\
        \bottomrule
    \end{tabular}
    \caption{Accuracy on the HardImageNet dataset with different segmentation setups. (top) GT masks, (middle) prompting with GT labels, (bottom) prompting with top-$k$ predictions of full image classifier without any ground truth labels or mask.}
    \label{tab:hin_setups}
\end{table}

\subsection{Supervised learning}

The setup of the experiments is described in the main text, where only a summary of the results was reported. Exhaustive results of all the base and fusion classifiers for all datasets are reported in Tables \ref{table:fungitastic:full} - \ref{tab:spaw:fullres}. 

\noindent\textbf{Stanford Dogs.}  Our experiments show that the context (\bg{} without shape) plays little role in breed identification on this dataset, see Table \ref{tab:stanfdogs:fullres}. 

\noindent\textbf{Resnet50 experiments on Spawrious.} These additional experiments use a LR of $ 10^{-5}$. As explained in Subsection \ref{subsec:spaw}, we set the training set \fg{}-\bg{} correlations to $ 97 \%$ to compare with the results in \cite{lynch2023spawrious}[Table 2]. The Resnet50 models are initialized with two sets of pretrained weights: from Timm and from torchvision. The results are recorded in Table \ref{table:spaw:resnetfull} 
\footnote{The results in Table \ref{table:spaw:resnetfull} are not directly comparable with Table \ref{tab:spaw:fullres} because \fg{}-\bg{} correlations are set differently. Also, the results slightly differ from those in the main text, where a sub-optimal learning rate for the \full{} model was used. This does not affect any of the conclusions}.

\noindent\textbf{Hard Imagenet with GT masks.} This setting provides an upper bound for the ``detection during recognition'' approach. The results are collected in Table \ref{tab:hingt:fullres}. The original test set has strong \fg{}-\bg{} correlations, and therefore \bg{} classifiers score very high by themselves and \fg{}+\bg{} performs best. On the long-tailed \bg{}s test set, \bg{} underperforms, but the \fg{} + \bg{} fusion still dominates. On the CT \bg{} test set all fusion models unsurprisingly underperform the\fg{}s.   

\begin{table}[tbh]
\centering
\setlength{\tabcolsep}{2.8pt}
\begin{tabular}{llll}
\toprule
 & Test & Test LT & Test CT \\
\midrule
\fg{} OWL  & $ 95.55 \% $ & $ 81.24 \%$ & $ 90.10 \%$ \\
\fg{} G-DINO  & $ 94.27\%$ & $ 78.32\%$ & $ 88.89\%$ \\
\midrule
\bg{} OWL  & $ 96.83\%$ & $ 80.27\%$ & $ 88.28\%$ \\
\bg{} G-DINO  & $ 95.20\%$ & $ 69.47\%$ & $ 83.84\%$ \\
\midrule
Fusion OWL & $ 98.03\%$ & $ 83.63\%$ & $ 91.31\%$ \\
Fusion G-DINO & $ 97.87\%$ & $ 80.09\%$ & $ 90.91\%$ \\
\bottomrule
\end{tabular}%
\caption{HardImageNet with automatic \fg{}-\bg{} generation and comparison of OWL vs GroundingDino for object proposal generation. Fusion consists of \full+\fg{}+\bg{}.}
\label{table:owl:gdino}
\end{table}

\paragraph{Hard Imagenet without GT prompt mask generation}
We also provide results of experiments with automatic \fg{}-\bg{} generation in the general setup of object recognition on images with many different objects on the Hard ImageNet dataset.

\noindent\textbf{Mask generation:} 
The \full{} classifier's top-$k$ predictions guide the segmentation prompt generation.
Let $ C $ be the number of classes. Each class $ i \in \overline{1,C} $ is described by a text prompt $ p_i $. 
For each sample, we consider the top-$ k $ predictions $ \{ i_j \}_{j=1}^k $of the \full{} model based on the confidence scores. These are the candidate labels for the final prediction and we prompt the detector with each of them, resulting in a \fg{} and \bg{} for each candidate. When the detctor output is empty, \fg{} and \bg{} are replaced by the full image.

\noindent\textbf{Fusion:} 
Unlike in the GT prompting/GT masks setup, the fusion now needs to take multiple candidates into account.

We tried applying the same fusion methods as in the main paper to each candidate individually, selecting the one with the highest fusion confidence as the output, but this approach does not outperform the \full{} baseline. This is likely caused by poor calibration of the fusion model output.

We provide an alternative hand-crafted fusion strategy which leads to positive results, but we do not claim it as a contribution and it may not transfer to other datasets:

Given an image $ x $ we aggregate logits for \fg{} and \bg{} predictions as follows. Using the top-$k$ prompts $ \{ p_{i_j} \}_{j=1}^k $ we generate $ x_{\text{FG}_{i_1}}, \dots  x_{\text{FG}_{i_k}} $ by equation \eqref{x_fg_bg_def},
\footnote{ If an image and prompt $ p_i $ fail to generate $ x_{\text{FG}_{i}} , x_{\text{BG}_{i}} $ using the detector then they default (fallback) to $ x_{\text{FG}_{i}} , x_{\text{BG}_{i}} := x $.}
which provide a list of output logits $ \Phi(x_{\text{FG}_{i_1}}), \dots  ,\Phi(x_{\text{FG}_{i_k}}) \in \mathbb{R}^C $ using the \fg{}-trained model $ \Phi $. From each of these vectors we record only the entry of the class of the prompt which generated it, i.e. entry $ i_j $ from $ \Phi(x_{\text{FG}_{i_j}}) $.  Selecting these, and following the same process for the \bg{} model $ \Psi $ we obtain aggregated logits
\begin{equation} \label{fg_bg_logits}
\begin{split}
z_{\text{FG}} &:= \left( \Phi(x_{\text{FG}_{i_1}})^{(i_1)}, \dots , \Phi(x_{\text{FG}_{i_k}})^{(i_k)} \right) \\
z_{\text{BG}} &:= \left( \Psi(x_{\text{BG}_{i_1}})^{(i_1)}, \dots , \Psi(x_{\text{BG}_{i_k}})^{(i_k)} \right)
\end{split}
\end{equation}
The \fg{} and \bg{} predictions are  $ i_{\ell} $ where $ \ell =  \argmax_j z_{\text{FG}}^{(j)} $, respectively $  \ell = \argmax_j z_{\text{BG}}^{(j)} $. 

For the fusions we combine \eqref{fg_bg_logits} with the original full image top-$k $ logits $ z_{\text{Full}} := (z_{i_1}, \dots z_{k_k}) $ by averaging $ 3 z_{Avg} := z_{\text{Full}} + z_{\text{FG}} + z_{\text{BG}} $ or $ 2 z_{Avg} := z_{\text{Full}} + z_{\text{FG}}  $.  The leads to prediction $ i_{\ell} $ if $ \ell =  \argmax_j z_{Avg}^{(j)} $. 

The results are provided in Table \ref{tab:hin_setups}.
An OWL vs Grounding DINO detector comparison is provided in Table \ref{table:owl:gdino}, showing the superiority of OWL in this setup.

\noindent\textbf{FungiTastic.} Since this dataset is highly unbalanced, we report the macro-averaged accuracy as the main metric. Due to some rare species, the number of present classes is smaller on the validation and test sets. The \emph{torchmetrics} implementation of the metrics, which we rely on in other experiments, does not account for such a scenario and the metric is implemented manually. Missing classes are removed before averaging the per-class accuracies on the validation and test sets.  

The results are summarized in Table \ref{table:fungitastic:full}. The highest mean accuracy is attained by the  \fg{} + \full{} combination. This shows that the \bg{} information (which is part of \full{}) is important for this dataset as well. 

The learnt weights of the weighted logits fusion model on the FungiTastic dataset are visualized in Figure \ref{fig:FungiWeights}.

\noindent\textbf{Segmentation ablation on Hard Imagent.}
In Table \ref{tab:hin_setups}, a comparison of the fully-automated segmentation to cheating segmentation setups is provided.
In the first set of experiments, ground truth segmentation masks from \cite{moayeri2022hard} are used to both train and evaluate all the models. 
In the second set of experiments, ground truth labels are used to create segmentation prompts during both training and evaluation. 
The last set of experiments is the standard fully-automatic setup which does not use any ground truth.

Surprisingly, the cheating setup with labels sometimes outperforms the ground truth masks. We hypothesize this can be attributed to the poor quality of the GT masks (coarse), compared to the outputs of SAM (clear shape).

\noindent\textbf{Per-class analysis on Hard ImageNet} where
\fg{}, \bg{} and fusion model performance is compared to \full{} on new test sets with strong \bg{} shift using GT masks is provided in Figure \ref{fig:hin_per_class}.
 Examples of \textbf{extreme overfitting to \bg{} on FungiTastic} are shown in Figure \ref{fig:fungi_bg_imp}.

\begin{table*}[htb] 
\centering 
\begin{tabular}{lllll}
\toprule
Dataset &  Val acc & Val avg acc & Test acc & Test avg acc \\
\midrule
\full{}  & $68.31\%_{\pm0.62}$ & $45.42\%_{\pm0.61}$ & $66.82\%_{\pm0.82}$ & $43.17\%_{\pm1.24}$ \\
\fg{}$_\text{C}$ &  $68.33\%_{\pm0.59}$ & $45.56\%_{\pm0.69}$ & $66.58\%_{\pm0.47}$ & $43.09\%_{\pm0.39}$ \\
\fg{}$_\text{M}$ &  $64.99\%_{\pm0.39}$ & $42.14\%_{\pm0.48}$ & $63.45\%_{\pm0.55}$ & $39.02\%_{\pm0.8}$ \\
\bg{}$_\text{S}$ & $45.26\%_{\pm2.92}$ & $24.73\%_{\pm2.42}$ & $43.1\%_{\pm2.42}$ & $23.76\%_{\pm2.07}$ \\
\bg{}$_\text{B}$ &  $21.79\%_{\pm0.34}$ & $10.94\%_{\pm0.38}$ & $19.84\%_{\pm0.3}$ & $10.77\%_{\pm0.47}$ \\
\midrule
\midrule
\fg{}$_\text{C}$ $\oplus$ \bg{}$_\text{S}$ \\
\midrule
\midrule
Max conf &  $68.08\%_{\pm0.22}$ & $43.54\%_{\pm1.5}$ & $66.23\%_{\pm0.26}$ & $41.55\%_{\pm1.02}$ \\
Max scaled conf & $68.0\%_{\pm0.46}$ & $43.37\%_{\pm0.9}$ & $66.06\%_{\pm0.27}$ & $41.26\%_{\pm0.52}$ \\
Threshold conf & $68.24\%_{\pm0.47}$ & $44.09\%_{\pm0.92}$ & $66.35\%_{\pm0.33}$ & $41.86\%_{\pm0.45}$ \\
TempScaled AvgPred &  $68.92\%_{\pm0.45}$ & $44.24\%_{\pm0.85}$ & $66.92\%_{\pm0.24}$ & $41.9\%_{\pm0.48}$ \\
TempScaled WeightedAvg &  $69.52\%_{\pm0.46}$ & $45.43\%_{\pm0.59}$ & $67.55\%_{\pm0.24}$ & $43.17\%_{\pm0.46}$ \\
Concatenate + FC layers &  $68.87\%_{\pm0.25}$ & $47.56\%_{\pm0.57}$ & $67.15\%_{\pm0.41}$ & $43.31\%_{\pm1.0}$ \\
WeightedLogitsComb & $70.47\%_{\pm0.17}$ & $\textbf{48.65}\%_{\pm0.52}$ & $68.58\%_{\pm0.39}$ & $45.65\%_{\pm0.35}$ \\
\midrule
Best - WeightedLogitsComb &    & $ 48.65\%_{\pm0.52}$ & & $45.65\%_{\pm0.35}$  \\
Oracle &  $74.19\%_{\pm0.49}$ & $50.77\%_{\pm0.69}$ & $72.46\%_{\pm0.41}$ & $48.88\%_{\pm0.94}$ \\
\midrule
\midrule
\fg{}$_\text{C}$ $\oplus $ \full{} \\
\midrule
\midrule
Max conf &  $71.47\%_{\pm0.43}$ & $47.34\%_{\pm1.04}$ & $69.64\%_{\pm0.35}$ & $45.17\%_{\pm0.88}$ \\
Max scaled conf &  $71.49\%_{\pm0.41}$ & $47.23\%_{\pm1.17}$ & $69.63\%_{\pm0.39}$ & $45.22\%_{\pm0.98}$ \\
Threshold conf &  $70.69\%_{\pm0.54}$ & $46.83\%_{\pm0.95}$ & $68.81\%_{\pm0.45}$ & $44.73\%_{\pm0.8}$ \\
TempScaled AvgPred & $71.96\%_{\pm0.38}$ & $48.27\%_{\pm0.96}$ & $70.18\%_{\pm0.28}$ & $45.85\%_{\pm0.81}$ \\
TempScaled WeightedAvg & $71.93\%_{\pm0.4}$ & $48.18\%_{\pm1.04}$ & $70.14\%_{\pm0.28}$ & $45.81\%_{\pm0.82}$ \\
Concatenate + FC layers &  $70.94\%_{\pm0.37}$ & $49.64\%_{\pm0.54}$ & $68.95\%_{\pm0.28}$ & $46.17\%_{\pm0.5}$ \\
WeightedLogitsComb &  $72.22\%_{\pm0.48}$ & $\textbf{51.15}\%_{\pm0.94}$ & $70.52\%_{\pm0.22}$ & $48.27\%_{\pm0.42}$ \\
\midrule
Best - WeightedLogitsComb &    & $ 51.15\%_{\pm0.94}$ & & $48.27\%_{\pm0.42}$  \\
Oracle & $97.69\%_{\pm2.82}$ &  $56.14\%_{\pm0.93}$ & $75.84\%_{\pm0.24}$ & $53.27\%_{\pm1.01}$ \\
\midrule 
\midrule
\full{} $\times 2$ Best - WLogitsComb &  & $51.18\%_{\pm0.28}$ &  & $48.54\%_{\pm0.43}$  \\
\midrule
\fg{}$_\text{C}$ $\times 2$ Best - WLogitsComb &  & $49.88\%_{\pm0.19}$ &  & $47.24\%_{\pm0.85}$  \\
\bottomrule
\bottomrule
\end{tabular}
\caption{FungiTastic results. Best corresponds to the best performing fusion on the validation set. Oracle prediction is correct if at least one of the fusion input's prediction is correct.}
\label{table:fungitastic:full}
\end{table*}

\begin{table*}[htb]
\centering 
\begin{tabular}{lllll}
\toprule
Dataset & Val acc & Test acc & Test Ad acc & Test Ct acc \\
\midrule
\full{}  & $98.25\%_{\pm0.15}$ & $97.33\%_{\pm0.13}$ & $81.33\%_{\pm1.01}$ & $90.51\%_{\pm0.9}$ \\
\fg{}$_\text{C}$  & $98.18\%_{\pm0.18}$ & $97.79\%_{\pm0.32}$ & $85.93\%_{\pm1.31}$ & $90.1\%_{\pm1.81}$ \\
\bg{}$_\text{S}$ & $98.4\%_{\pm0.06}$ & $97.84\%_{\pm0.55}$ & $79.73\%_{\pm1.86}$ & $73.94\%_{\pm2.71}$ \\
\midrule
\midrule
\fg{}$_\text{C}$ $\oplus$ \bg{}$_\text{S}$ \\
\midrule
\midrule
Max conf  & $98.75\%_{\pm0.09}$ & $98.77\%_{\pm0.22}$ & $90.0\%_{\pm0.8}$ & $91.11\%_{\pm3.38}$ \\
Max scaled conf  & $98.84\%_{\pm0.04}$ & $98.88\%_{\pm0.31}$ & $89.91\%_{\pm0.48}$ & $91.31\%_{\pm2.63}$ \\
Threshold conf  & $98.27\%_{\pm0.15}$ & $98.11\%_{\pm0.24}$ & $86.99\%_{\pm0.8}$ & $90.91\%_{\pm1.01}$ \\
TempScaled AvgPred  & $98.85\%_{\pm0.05}$ & $98.91\%_{\pm0.29}$ & $90.44\%_{\pm0.86}$ & $91.31\%_{\pm1.53}$ \\
TempScaled WeightedAvg  & $98.74\%_{\pm0.12}$ & $98.64\%_{\pm0.36}$ & $87.35\%_{\pm2.33}$ & $85.86\%_{\pm5.76}$ \\
Concatenate + FC layers  & $98.94\%_{\pm0.02}$ & $98.99\%_{\pm0.2}$ & $90.0\%_{\pm1.31}$ & $92.12\%_{\pm1.32}$ \\
WeightedLogitsComb & $98.85\%_{\pm0.14}$ & $98.93\%_{\pm0.33}$ & $90.44\%_{\pm0.4}$ & $90.51\%_{\pm1.69}$ \\
\midrule 
Best - Concat+ FC  &  $98.94\%_{\pm0.02}$ & $98.99\%_{\pm0.2}$ & $90.0\%_{\pm1.31}$ & $92.12\%_{\pm1.32}$ \\
Oracle & $99.36\%_{\pm0.08}$ & $99.36\%_{\pm0.11}$ & $94.07\%_{\pm1.2}$ & $96.77\%_{\pm1.94}$ \\
\midrule
\midrule
\fg{}$_\text{C}$ $\oplus $ \full{} \\
\midrule
\midrule
Max conf  & $98.69\%_{\pm0.1}$ & $98.72\%_{\pm0.15}$ & $89.12\%_{\pm1.2}$ & $90.1\%_{\pm1.32}$ \\
Max scaled conf  & $98.82\%_{\pm0.18}$ & $98.8\%_{\pm0.16}$ & $89.12\%_{\pm1.35}$ & $90.71\%_{\pm0.85}$ \\
Threshold conf  & $98.26\%_{\pm0.15}$ & $98.11\%_{\pm0.3}$ & $87.17\%_{\pm0.54}$ & $89.9\%_{\pm1.24}$ \\
TempScaled AvgPred  & $98.84\%_{\pm0.16}$ & $98.96\%_{\pm0.15}$ & $89.38\%_{\pm1.08}$ & $90.91\%_{\pm1.01}$ \\
TempScaled WeightedAvg  & $98.74\%_{\pm0.07}$ & $98.61\%_{\pm0.31}$ & $87.96\%_{\pm0.58}$ & $90.91\%_{\pm1.01}$ \\
Concatenate + FC layers  & $98.85\%_{\pm0.06}$ & $98.85\%_{\pm0.15}$ & $88.76\%_{\pm1.27}$ & $90.1\%_{\pm1.11}$ \\
WeightedLogitsComb  & $98.82\%_{\pm0.05}$ & $99.01\%_{\pm0.28}$ & $89.38\%_{\pm1.25}$ & $90.71\%_{\pm0.85}$ \\
\midrule 
Best - Concat+ FC &  $98.85\%_{\pm0.06}$ & $98.85\%_{\pm0.15}$  & $88.76\%_{\pm1.27}$ & $90.1\%_{\pm1.11}$  \\
Oracle  & $99.32\%_{\pm0.05}$ & $99.47\%_{\pm0.09}$ & $92.57\%_{\pm0.48}$ & $92.12\%_{\pm1.11}$ \\
\midrule
\midrule
\full{} $ \times 2$ Best - TempScaled WAvg   &  $98.45\%_{\pm0.2}$ & $97.51\%_{\pm0.2}$ & $82.89\%_{\pm0.26}$ & $90.91\%_{\pm1.01}$ \\ 
\midrule
\fg{}$_\text{C}$ $ \times 2$ Best  - TempScaled WAvg &  $98.52\%_{\pm0.18}$ & $97.91\%_{\pm0.28}$ & $86.58\%_{\pm2.55}$ & $89.9\%_{\pm1.01}$ \\
\bottomrule
\bottomrule
\end{tabular}
\caption{HardImageNet results using OWLv2 generated masks using GT prompts. Best corresponds to the best performing fusion on the validation set. Oracle prediction is correct if at least one of the fusion input's prediction is correct.}
\label{tab:hinowlgt:fullres}
\end{table*}

\begin{table*}[htb]
\centering 
\begin{tabular}{lllll}
\toprule
Dataset &  Val acc & Test acc & Test LT acc & Test CT acc \\
\midrule
\full{}  & $98.25\%_{\pm0.15}$ & $97.33\%_{\pm0.13}$ & $81.33\%_{\pm1.01}$ & $90.51\%_{\pm0.9}$ \\
\fg{}$_\text{C}$  & $98.19\%_{\pm0.13}$ & $97.63\%_{\pm0.3}$ & $87.08\%_{\pm1.34}$ & $94.55\%_{\pm1.15}$ \\
\fg{}$_\text{M}$  & $95.6\%_{\pm0.22}$ & $95.39\%_{\pm1.15}$ & $85.4\%_{\pm1.77}$ & $95.56\%_{\pm1.53}$ \\
\bg{}$_\text{S}$ & $97.5\%_{\pm0.1}$ & $95.01\%_{\pm0.2}$ & $78.94\%_{\pm1.52}$ & $67.88\%_{\pm3.53}$ \\
\bg{}$_B$  & $94.02\%_{\pm0.31}$ & $91.76\%_{\pm0.49}$ & $56.64\%_{\pm2.1}$ & $24.24\%_{\pm2.77}$ \\
\midrule
\midrule
\fg{}$_\text{C}$ $\oplus$ \bg{}$_\text{S}$ \\
\midrule
\midrule
Max conf  & $98.89\%_{\pm0.16}$ & $98.29\%_{\pm0.2}$ & $88.58\%_{\pm1.01}$ & $90.91\%_{\pm2.26}$ \\
Max scaled conf  & $99.0\%_{\pm0.16}$ & $98.37\%_{\pm0.26}$ & $88.23\%_{\pm0.92}$ & $90.51\%_{\pm3.16}$ \\
Threshold conf & $98.26\%_{\pm0.1}$ & $97.63\%_{\pm0.37}$ & $87.7\%_{\pm1.75}$ & $94.14\%_{\pm1.11}$ \\
TempScaled AvgPred  & $99.06\%_{\pm0.2}$ & $98.35\%_{\pm0.26}$ & $88.5\%_{\pm1.04}$ & $91.11\%_{\pm3.15}$ \\
TempScaled WeightedAvg  & $98.82\%_{\pm0.25}$ & $98.16\%_{\pm0.2}$ & $88.05\%_{\pm1.68}$ & $90.51\%_{\pm4.66}$ \\
Concatenate + FC layers  & $99.08\%_{\pm0.23}$ & $98.35\%_{\pm0.12}$ & $89.2\%_{\pm1.31}$ & $90.1\%_{\pm2.52}$ \\
WeightedLogitsComb  & $ \textbf{99.13}\%_{\pm0.19}$ & $98.45\%_{\pm0.18}$ & $89.56\%_{\pm1.35}$ & $90.3\%_{\pm2.91}$ \\
\midrule 
Best - WeightedLogitsComb& $99.13\%_{\pm0.19}$ & $98.45\%_{\pm0.18}$ & $89.56\%_{\pm1.35}$ & $90.3\%_{\pm2.91}$ \\
Oracle & $99.46\%_{\pm0.18}$ & $99.23\%_{\pm0.15}$ & $93.27\%_{\pm0.48}$ & $95.56\%_{\pm1.83}$ \\
\midrule
\midrule
\fg{}$_\text{C}$ $\oplus $ \full{} \\
\midrule
\midrule
Max conf  & $98.8\%_{\pm0.14}$ & $98.27\%_{\pm0.35}$ & $88.76\%_{\pm0.92}$ & $93.54\%_{\pm2.63}$ \\
Max scaled conf  & $98.87\%_{\pm0.18}$ & $98.4\%_{\pm0.34}$ & $88.85\%_{\pm0.79}$ & $93.54\%_{\pm2.91}$ \\
Threshold conf  & $98.26\%_{\pm0.13}$ & $97.63\%_{\pm0.37}$ & $88.14\%_{\pm1.45}$ & $94.75\%_{\pm1.5}$ \\
TempScaled AvgPred  & $98.88\%_{\pm0.22}$ & $98.48\%_{\pm0.24}$ & $89.12\%_{\pm0.92}$ & $93.33\%_{\pm2.91}$ \\
TempScaled WeightedAvg  & $98.81\%_{\pm0.14}$ & $98.29\%_{\pm0.22}$ & $88.76\%_{\pm0.67}$ & $93.33\%_{\pm2.73}$ \\
Concatenate + FC layers  & $\textbf{99.01}\%_{\pm0.14}$ & $98.48\%_{\pm0.24}$ & $88.94\%_{\pm0.83}$ & $93.13\%_{\pm1.94}$ \\
WeightedLogitsComb & $99.0\%_{\pm0.14}$ & $98.48\%_{\pm0.2}$ & $89.65\%_{\pm0.86}$ & $93.13\%_{\pm2.41}$ \\
\midrule 
Best - Concat+ FC &   $99.01\%_{\pm0.14}$ & $98.48\%_{\pm0.24}$ & $88.94\%_{\pm0.83}$ & $93.13\%_{\pm1.94}$ \\
Oracle  & $99.46\%_{\pm0.1}$ & $99.36\%_{\pm0.26}$ & $93.27\%_{\pm0.58}$ & $96.16\%_{\pm1.94}$ \\
\midrule
\midrule
\full{} $ \times 2$ Best - TempScaled WAvg  &  $98.45\%_{\pm0.2}$ & $97.51\%_{\pm0.2}$ & $82.89\%_{\pm0.26}$ & $90.91\%_{\pm1.01}$ \\ 
\midrule
\fg{}$_\text{C}$ $ \times 2$ Best - Concat + FC  &   $98.43\%_{\pm0.09}$ & $97.96\%_{\pm0.34}$ & $87.17\%_{\pm0.77}$ & $94.95\%_{\pm1.01}$ \\
\bottomrule
\bottomrule
\end{tabular}
\caption{HardImageNet results using coarse ground truth masks provided by the original dataset authors. Best corresponds to the best performing fusion on the validation set. Oracle prediction is correct if at least one of the fusion input's prediction is correct.}
\label{tab:hingt:fullres}
\end{table*}

\begin{table*}[htb]
\centering 
\begin{tabular}{llll}
\toprule
Dataset & Train acc & Val acc & Test acc \\
\midrule
\full{} & $98.77\%_{\pm1.01}$ & $90.01\%_{\pm0.42}$ & $90.28\%_{\pm0.32}$ \\
\fg{}$_\text{C}$ & $97.28\%_{\pm1.56}$ & $90.89\%_{\pm0.36}$ & $91.25\%_{\pm0.25}$ \\
\fg{}$_\text{M}$ & $97.24\%_{\pm1.68}$ & $89.2\%_{\pm0.41}$ & $89.17\%_{\pm0.35}$ \\
\bg{}$_\text{S}$ & $97.36\%_{\pm2.53}$ & $51.8\%_{\pm0.74}$ & $51.34\%_{\pm0.94}$ \\
\bg{}$_B$ & $96.99\%_{\pm2.72}$ & $8.0\%_{\pm1.12}$ & $7.76\%_{\pm1.46}$ \\
\midrule
\midrule
\fg{}$_\text{C}$ $\oplus$ \bg{}$_\text{S}$ \\
\midrule
\midrule
Max conf  & $98.75\%_{\pm1.27}$ & $83.54\%_{\pm4.1}$ & $83.78\%_{\pm4.07}$ \\
Max scaled conf & $98.27\%_{\pm1.29}$ & $89.98\%_{\pm0.35}$ & $90.51\%_{\pm0.25}$ \\
Threshold conf & $97.58\%_{\pm1.46}$ & $90.59\%_{\pm0.33}$ & $90.94\%_{\pm0.21}$ \\
TempScaled AvgPred & $98.46\%_{\pm1.16}$ & $90.0\%_{\pm0.28}$ & $90.59\%_{\pm0.27}$ \\
TempScaled WeightedAvg & $97.45\%_{\pm1.53}$ & $90.89\%_{\pm0.42}$ & $91.27\%_{\pm0.29}$ \\
Concatenate + FC layers & $99.65\%_{\pm0.31}$ & $80.16\%_{\pm4.54}$ & $79.77\%_{\pm4.88}$ \\
WeightedLogitsComb & $99.07\%_{\pm0.83}$ & $87.29\%_{\pm2.09}$ & $87.21\%_{\pm2.1}$ \\
\midrule 
Best - TempScaled WAvg&  & $90.89\%_{\pm0.42}$ & $91.27\%_{\pm0.29}$ \\
Oracle & $99.48\%_{\pm0.58}$ & $92.79\%_{\pm0.31}$ & $93.29\%_{\pm0.21}$ \\
\midrule
\midrule
\fg{}$_\text{C}$ $\oplus $ \full{}  \\
\midrule
\midrule
Max conf & $98.76\%_{\pm1.04}$ & $91.18\%_{\pm0.54}$ & $91.69\%_{\pm0.62}$ \\
Max scaled conf & $98.64\%_{\pm0.88}$ & $91.43\%_{\pm0.35}$ & $92.16\%_{\pm0.22}$ \\
Threshold conf & $97.56\%_{\pm1.38}$ & $91.08\%_{\pm0.27}$ & $91.66\%_{\pm0.25}$ \\
TempScaled AvgPred & $98.67\%_{\pm0.88}$ & $91.41\%_{\pm0.33}$ & $92.15\%_{\pm0.23}$ \\
TempScaled WeightedAvg & $98.27\%_{\pm1.07}$ & $91.34\%_{\pm0.41}$ & $92.01\%_{\pm0.33}$ \\
Concatenate + FC layers & $99.2\%_{\pm0.66}$ & $91.37\%_{\pm0.32}$ & $91.63\%_{\pm0.44}$ \\
WeightedLogitsComb & $98.85\%_{\pm0.94}$ & $91.34\%_{\pm0.26}$ & $92.06\%_{\pm0.31}$ \\
\midrule 
Best - Max scaled conf  &  & $91.43\%_{\pm0.35}$ & $92.16\%_{\pm0.22}$ \\
Oracle & $99.17\%_{\pm0.73}$ & $93.9\%_{\pm0.4}$ & $94.47\%_{\pm0.3}$ \\
\midrule
\midrule
\full{} $ \times 2$ Best - Concat + FC  &  & $91.07\%_{\pm0.23}$ & $91.13\%_{\pm0.17}$ \\
\midrule
\fg{}$_\text{C}$ $ \times 2$ Best - Concat + FC  &  &   $91.65\%_{\pm0.42}$ & $91.9\%_{\pm0.03}$ \\
\bottomrule
\bottomrule
\end{tabular}
\caption{Stanford dogs. Best corresponds to the best performing fusion on the validation set. Oracle prediction is correct if at least one of the fusion input's prediction is correct.} 
\label{tab:stanfdogs:fullres}
\end{table*}

\begin{table*}[htb]
\centering 
\begin{tabular}{lll}
\toprule
Dataset & Test acc & Test clean \\
\midrule
\full{} & $82.35$ & $92.01$ \\
\fg{}$_\text{C}$ & $85.56$ & $91.99$ \\
\bg{}$_\text{S}$ & $73.24$ & $81.28$ \\
\midrule
\midrule
\fg{}$_\text{C}$ $\oplus$ \bg{}$_\text{S}$ \\
\midrule
\midrule
Max conf  & $86.39$ & $93.02$ \\
Max scaled conf & $86.57$ & $93.19$ \\
Threshold conf & $86.57$ & $93.18$ \\
TempScaled AvgPred & $86.77$ & $93.32$ \\
TempScaled WeightedAvg & $86.93$ & $93.33$ \\
Concatenate + FC layers & $86.38$ & $92.76$ \\
WeightedLogitsComb & $87.13$ & $93.30$ \\
\midrule 
Best - WeightedLogitsComb & $87.13$ & $93.30$ \\
\midrule
\midrule
\fg{}$_\text{C}$ $\oplus $ \full{}  \\
\midrule
\midrule
Max conf & $86.22$ & $93.50$ \\
Max scaled conf & $86.38$ & $93.48$ \\
Threshold conf & $86.38$ & $93.48$ \\
TempScaled AvgPred & $86.58$ & $93.60$ \\
TempScaled WeightedAvg & $86.41$ & $93.57$ \\
Concatenate + FC layers & $86.00$ & $92.77$ \\
WeightedLogitsComb & $87.04$ & $93.76$ \\
\midrule 
Best - WeightedLogitsComb & $87.04$ & $93.76$ \\
\bottomrule
\bottomrule
\end{tabular}
\caption{ImageNet results using OWLv2 generated masks using GT prompts. Best corresponds to the best performing fusion on the validation set.} 
\label{tab:imagenet:fullres}
\end{table*}

\begin{table*}[bth]
\centering 
\begin{tabular}{llll}
\toprule
Dataset & Train acc & Val acc & Test acc \\
\midrule
\full{} & $100.0\%_{\pm0.01}$ & $100.0\%_{\pm0.0}$ & $43.2\%_{\pm9.74}$ \\
\fg{}$_\text{C}$ & $99.96\%_{\pm0.07}$ & $99.91\%_{\pm0.05}$ & $91.31\%_{\pm3.45}$ \\
\fg{}$_\text{M}$ & $99.96\%_{\pm0.06}$ & $99.7\%_{\pm0.08}$ & $95.28\%_{\pm1.47}$ \\
\bg{}$_S$ & $100.0\%_{\pm0.0}$ & $99.83\%_{\pm0.04}$ & $2.62\%_{\pm0.8}$ \\
\bg{}$_B$ & $99.94\%_{\pm0.06}$ & $99.18\%_{\pm0.12}$ & $0.18\%_{\pm0.08}$ \\
\midrule
\midrule
\fg{}$_\text{C}$ $\oplus$ \bg{}$_S$ \\
\midrule
\midrule
Max conf & $100.0\%_{\pm0.0}$ & $99.98\%_{\pm0.02}$ & $25.9\%_{\pm20.5}$ \\
Max scaled conf & $100.0\%_{\pm0.0}$ & $100.0\%_{\pm0.0}$ & $66.15\%_{\pm28.27}$ \\
Threshold conf & $99.96\%_{\pm0.07}$ & $99.91\%_{\pm0.05}$ & $91.25\%_{\pm3.47}$ \\
TempScaled AvgPred & $100.0\%_{\pm0.0}$ & $100.0\%_{\pm0.0}$ & $65.81\%_{\pm27.59}$ \\
TempScaled WeightedAvg & $100.0\%_{\pm0.0}$ & $99.99\%_{\pm0.02}$ & $70.39\%_{\pm38.02}$ \\
Concatenate + FC layers & $100.0\%_{\pm0.0}$ & $99.97\%_{\pm0.04}$ & $49.68\%_{\pm11.85}$ \\
WeightedLogitsComb & $100.0\%_{\pm0.0}$ & $99.98\%_{\pm0.02}$ & $27.71\%_{\pm18.67}$ \\
\midrule 
Best & & $100.0\%_{\pm0.0}$ & 65.81-66.15\%  \\
Oracle & $100.0\%_{\pm0.0}$ & $100.0\%_{\pm0.0}$ & $91.32\%_{\pm3.45}$ \\
\midrule
\midrule
\fg{}$_\text{C}$ $\oplus $ \full{} \\
\midrule
\midrule
Max conf & $100.0\%_{\pm0.0}$ & $100.0\%_{\pm0.0}$ & $76.66\%_{\pm7.28}$ \\
Max scaled conf & $100.0\%_{\pm0.0}$ & $100.0\%_{\pm0.0}$ & $51.86\%_{\pm16.57}$ \\
Threshold conf & $99.96\%_{\pm0.07}$ & $99.91\%_{\pm0.05}$ & $91.26\%_{\pm3.47}$ \\
TempScaled AvgPred & $100.0\%_{\pm0.0}$ & $100.0\%_{\pm0.0}$ & $52.02\%_{\pm16.71}$ \\
TempScaled WeightedAvg & $100.0\%_{\pm0.01}$ & $100.0\%_{\pm0.0}$ & $43.2\%_{\pm9.74}$ \\
Concatenate + FC layers & $100.0\%_{\pm0.0}$ & $100.0\%_{\pm0.0}$ & $68.41\%_{\pm14.15}$ \\
WeightedLogitsComb & $100.0\%_{\pm0.0}$ & $100.0\%_{\pm0.0}$ & $76.95\%_{\pm6.83}$ \\
\midrule 
Best & & $100.0\%_{\pm0.0}$ & 43.06-76.95\%\\
Oracle & $100.0\%_{\pm0.0}$ & $100.0\%_{\pm0.0}$ & $91.57\%_{\pm3.47}$ \\
\midrule
\midrule
\full{} $ \times 2$ Best   &  & $100.0\%_{\pm0.0}$ &  40-45\%\\
\midrule
\fg{}$_\text{C}$ $ \times 2$ Best - Concat + FC  &  &   $100.0\%_{\pm0.0}$ & $94.62\%_{\pm1.65}$ \\
\bottomrule
\bottomrule
\end{tabular}
\caption{Result of ConvNeXt models on the Spawrious dataset. Best corresponds to the best performing fusion on the validation set. Oracle prediction is correct if at least one of the fusion input's prediction is correct. 
} 
\label{tab:spaw:fullres}
\end{table*}

\begin{table*}[bth]
\centering
\begin{tabular}{llll}
\toprule
Dataset & Train acc & Val acc & Test acc \\
\midrule
\multicolumn{4}{c}{Timm Resnet50} \\
\midrule
\full{} & $99.76\%_{\pm0.06}$ & $99.3\%_{\pm0.19}$ & $87.38\%_{\pm0.71}$ \\
\midrule
$\text{\fg{}}_\text{C}$ undistorted & $99.46\%_{\pm0.05}$ & $99.12\%_{\pm0.1}$ & $95.02\%_{\pm1.04}$ \\
$\text{\fg{}}_\text{C}$ distorted & $99.06\%_{\pm0.08}$ & $98.34\%_{\pm0.22}$ & $94.26\%_{\pm0.95}$ \\
$\text{\fg{}}_\text{M}$ distorted & $98.81\%_{\pm0.03}$ & $98.36\%_{\pm0.11}$ & $94.83\%_{\pm0.29}$ \\
$\text{\fg{}}_\text{M}$ undistorted & $99.06\%_{\pm0.19}$ & $98.69\%_{\pm0.17}$ & $95.42\%_{\pm0.52}$ \\
\midrule
$\text{\bg{}}_\text{S}$ & $98.66\%_{\pm0.12}$ & $97.54\%_{\pm0.12}$ & $24.26\%_{\pm4.39}$ \\
$\text{\bg{}}_\text{B}$ & $95.67\%_{\pm0.13}$ & $94.89\%_{\pm0.27}$ & $0.72\%_{\pm0.14}$ \\
\midrule
\fg{}$\oplus_{\text{R}}$\bg{}  & $99.79\%_{\pm0.05}$ & $99.55\%_{\pm0.04}$ & $91.37\%_{\pm0.8}$ \\
\midrule
\multicolumn{4}{c}{Torchvision Resnet50} \\
\midrule
\full{} & $100.0\%_{\pm0.0}$ & $99.85\%_{\pm0.04}$ & $71.35\%_{\pm3.72}$ \\
\midrule
$\text{\fg{}}_\text{C}$ undistorted & $100.0\%_{\pm0.0}$ & $99.83\%_{\pm0.05}$ & $95.0\%_{\pm0.63}$ \\
$\text{\fg{}}_\text{C}$ distorted & $99.99\%_{\pm0.01}$ & $99.61\%_{\pm0.08}$ & $94.97\%_{\pm0.44}$ \\
$\text{\fg{}}_\text{M}$ distorted & $100.0\%_{\pm0.0}$ & $99.42\%_{\pm0.08}$ & $95.22\%_{\pm0.12}$ \\
$\text{\fg{}}_\text{M}$ undistorted & $100.0\%_{\pm0.0}$ & $99.58\%_{\pm0.05}$ & $95.59\%_{\pm0.25}$ \\
\midrule
$\text{\bg{}}_\text{S}$ & $100.0\%_{\pm0.0}$ & $99.21\%_{\pm0.16}$ & $8.9\%_{\pm1.38}$ \\
$\text{\bg{}}_\text{B}$ & $99.76\%_{\pm0.3}$ & $96.94\%_{\pm0.08}$ & $0.36\%_{\pm0.03}$ \\
\midrule
\fg{}$\oplus_{\text{R}}$\bg{} & $100.0\%_{\pm0.0}$ & $99.91\%_{\pm0.06}$ & $86.78\%_{\pm3.95}$ \\
\bottomrule
\end{tabular}
\caption{Spawrious. Resnet50 models, same architecture, with two different initializations (timm and torchvision). The Timm pretrained checkpoints are significantly more robust. 
} 
\label{table:spaw:resnetfull}
\end{table*}

\subsection{Vision-Language Models}
A comparison of CLIP-B (openai/clip-vit-base-patch32), BioCLIP (imageomics/bioclip), CLIP-L (openai/clip-vit-large-patch14) and SigLIP2-SO (timm/ViT-SO400M-16-SigLIP2-256) are presented in
Table \ref{tab:vlm_comp_full}.
The results show that \strr{} with \fg{} $\oplus_{\text{max}}$ \full{} fusion consistently improves performance across all test sets. The only exceptions are the `rare' CounterAnimal test set for some of the models and BioCLIP, which was trained for very different domains than most of the datasets. On its target domain, the FungiTastic, its performance doubles with \strr{}.
The biggest gains are achieved for the smallest CLIP-B model, whose performance is significantly lower on average.

\end{document}